\PassOptionsToPackage{numbers,square,sort&compress}{natbib}
\PassOptionsToPackage{table,dvipsnames}{xcolor}
\documentclass[]{style/company_light} % Switch to style/company_border for border variant
\usepackage[utf8]{inputenc}
\usepackage{hyperref}
\usepackage{url}
\usepackage{booktabs}
\usepackage{amsfonts}
\usepackage{amsmath}
\usepackage{nicefrac}
\usepackage{microtype}
\usepackage{graphicx}
\usepackage{float}
\usepackage{subcaption}
\usepackage{multirow}
\usepackage{enumerate}
\usepackage{enumitem}
\usepackage[most]{tcolorbox}
\usepackage{adjustbox}
\usepackage{array}
\usepackage{amssymb}
\usepackage{tabularx}
\usepackage{longtable}
\usepackage{listings}
\usepackage{needspace}

\makeatletter
\def\fps@figure{htbp}
\def\fps@table{htbp}
\makeatother

\newcommand{\compactfloatlayout}{%
  \renewcommand{\topfraction}{0.92}%
  \renewcommand{\bottomfraction}{0.88}%
  \renewcommand{\textfraction}{0.08}%
  \renewcommand{\floatpagefraction}{0.85}%
  \setcounter{topnumber}{3}%
  \setcounter{bottomnumber}{2}%
  \setcounter{totalnumber}{4}%
  \setlength{\textfloatsep}{10pt plus 2pt minus 2pt}%
  \setlength{\intextsep}{10pt plus 2pt minus 2pt}%
  \setlength{\floatsep}{8pt plus 2pt minus 2pt}%
  \interlinepenalty=0\relax
  \clubpenalty=10000\relax
  \widowpenalty=10000\relax
  \raggedbottom
}

\definecolor{codekeyword}{HTML}{754290}
\definecolor{codecall}{HTML}{315F85}
\definecolor{codestring}{HTML}{9A4B17}
\definecolor{codecomment}{HTML}{526963}
\definecolor{codebackground}{HTML}{F1F5FA}
\lstdefinestyle{experience}{
  language=Python,
  basicstyle=\fontfamily{pcr}\fontsize{7.5}{9}\selectfont,
  keywordstyle=\color{codekeyword}\bfseries,
  commentstyle=\color{codecomment},
  stringstyle=\color{codestring},
  emph={get_observation,get_instruction,step,move_ee,move_both,pick_from_above},
  emphstyle=\color{codecall},
  backgroundcolor=\color{codebackground},
  columns=fullflexible,keepspaces=true,showstringspaces=false,
  breaklines=true,breakatwhitespace=false,breakindent=1em,
  frame=single,framerule=0pt,framesep=4pt,
  xleftmargin=4pt,xrightmargin=4pt,aboveskip=5pt,belowskip=5pt
}

\NewDocumentCommand{\todo}
{ mO{} }{\textcolor{magenta}{\textsuperscript{\textit{TODO}}\textsf{\textbf{\small[#1]}}}}
\title{\fontsize{18pt}{21pt}\selectfont\baselineskip=21pt\relax
EmbodiedEvo: Self-Evolving Coding Agent For Physical Intelligence\par}

\author{
    \footnotesize Wenyi Li$^{\dagger}$, Yizhe Chi, Youjie Zheng,  Chunyu Liu, Xiaoqiu Wang, Mingju Gao, Zheng Jiang, Deyao Hong, Calvin Xiao,  Kaisen Yang$^{\ddagger}$, Qinhuai Na$^{\ddagger}$
}
\renewcommand\affiliationformat[2][]{\makebox[\linewidth][c]{\small\bfseries #2}}
\affiliation{Navers Lab, Einsia.AI,\quad\quad Tsinghua University}
\renewcommand\contributionformat[2][]{%
  \vskip 0.15cm
  \makebox[\linewidth][c]{\footnotesize\color{gray}#2}%
}
\contribution{$^{\dagger}$Project Lead\quad$^{\ddagger}$Corresponding Authors}

\abstract{%
Coding agents can construct executable programs and revise them through feedback, offering a foundation for embodied control without target-task fine-tuning. Through a programmatic robot interface, the agent observes the scene, executes control programs, and adapts them to the resulting physical state, serving as the closed-loop policy. However, this within-episode adaptation does not automatically preserve experience across scenes. Our trajectory analysis shows that agents spend interaction and inference budgets establishing effective operations and identifying their preconditions, while fresh agents cannot reuse these discoveries.
This motivates \textbf{EmbodiedEvo}, a self-evolving coding agent that distills physical feedback into reusable code Skills and diagnostic Lessons. Experience is acquired on one Playground scene per task, then frozen for independent agents in unseen scenes, without updating model weights.
On RoboDojo-Sim, EmbodiedEvo achieves an average score of 44.48 and a success rate of 37.82\%, surpassing all officially verified results. Paired tests show the largest gains on memory-dependent tasks and successful reuse of coordinated procedures, while fine contact control remains difficult. Experience from stronger models also improves weaker executors: Astra's experience raises Sol and Luna's success rates by 8.86 and 5.50 percentage points over no experience. Compared with raw interaction history, Skills and Lessons achieve comparable task performance at 53.0\% lower execution cost.
}

\date{October 13, 2026}
\metadata[Homepage]{\url{https://lab.einsia.ai/embodiedevo}}
\correspondence{\email{liwenyi@einsia.ai}, \email{kaisenyang@einsia.ai},  \email{nana@einsia.ai}}

\begin{document}

\maketitle

\begin{figure}[!htb]
    \centering
    \includegraphics[width=1.0\linewidth]{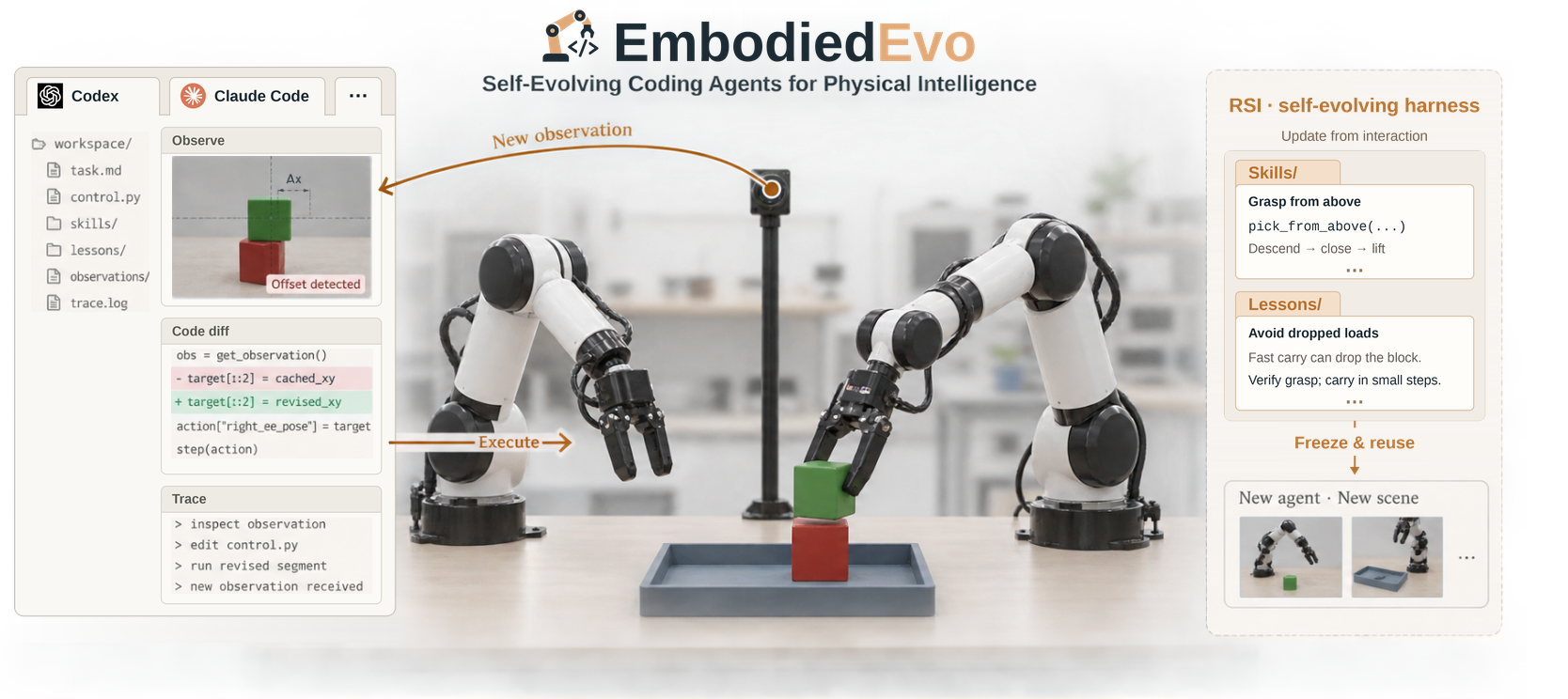}
    \caption{\textbf{Overview of EmbodiedEvo.} During task-specific Playground interaction, a coding agent develops control programs and revises reusable Skills and Lessons through physical feedback. The resulting harness is frozen and inherited by fresh Test agents on new scenes.}
    \label{fig:overview}
\end{figure}
\FloatBarrier

\compactfloatlayout
\clearpage
\section{Introduction}
\label{sec:introduction}

% \begin{center}
%     \small\rmfamily\mdseries\itshape
%     % Source: Intelligence Without Reason (1991), Section 5.3.
%     % https://people.csail.mit.edu/brooks/papers/BrooksIJCAI91.pdf
%     ``Intelligence is determined by the dynamics of interaction with the world.''\\[3pt]
%     \makebox[\linewidth][r]{\upshape --- Rodney A. Brooks}
% \end{center}

Coding agents offer a promising foundation for embodied control by combining program construction, tool use, and feedback-driven debugging~\citep{fu2026capx,liang2022codeaspolicies,shen2026embodiedswe,song2026embodiedrsi,ding2026zetta}. Through a programmatic robot interface, these capabilities allow an agent to translate task goals into executable control procedures, inspect their physical outcomes, and revise the procedures as execution unfolds. 
The coding agent thus serves as the closed-loop policy, deciding what to observe, which programs to execute, and how to respond to feedback. This approach already shows promising zero-shot performance, where frontier agentic models could achieve a success rate comparable to VLA or WAM systems trained on target-similar scenes but in a zero-shot manner~\cite{zhang2026unexpected}. This highlights the potential of general-purpose coding capabilities for physical task execution.

These zero-shot results, however, represent only a starting point for what coding agents can achieve in embodied control. In the zero-shot setting, an agent begins operating an unfamiliar embodiment without an opportunity to first learn from physical interaction. Understanding the task and the robot interface does not establish how to reliably realize an intended action: a valid command may move the gripper to the desired pose without securing the object. The agent must therefore discover effective operations and their preconditions during execution, spending part of its task budget familiarizing itself with the embodiment. While VLA and WAM systems acquire relevant experience through training on target-similar scenes, zero-shot coding agents undertake this adaptation while attempting to complete the task. This raises a natural question: can a coding agent improve its performance in unseen scenes if it is first given an opportunity to explore and acquire experience with the embodiment?

To this end, we introduce \textbf{EmbodiedEvo}, a self-evolving coding agent that acquires reusable physical experience through dedicated interaction with its embodiment. In a Playground containing one scene per task, the agent explores, inspects physical feedback, and refines two complementary forms of experience: executable Skills that capture how to perform operations, and diagnostic Lessons that record their applicability conditions, observed failure modes, and recovery guidance. The resulting experience is then frozen and provided to fresh agents executing the same task in unseen layouts, without updating model weights. These agents reuse the acquired experience while continuing to adapt their control programs through closed-loop interaction.

Experiments on RoboDojo~\citep{zhang2026unexpected} simulation tasks show that inherited experience improves subsequent execution. With task descriptions and wiki recipes, Astra/xhigh reaches 37.82\% Average success rate over 50 Test scenes per task, compared with the published Astra result of 22.48\% . Paired tests hold the executor and scenes fixed, raising success from 8.50\% to 30.17\% with task descriptions alone. The largest gains occur on memory-dependent tasks, alongside successful reuse of coordinated procedures such as block stacking; fine insertion and liquid handling remain difficult. Experience also transfers across models: Astra's Skills and Lessons raise Sol and Luna's success rates by 8.86 and 5.50 percentage points over no experience. These results distinguish the ability to produce useful control knowledge from the ability to apply it.

Our contributions are:
\begin{itemize}[leftmargin=*,itemsep=3pt,topsep=3pt]
    \item \textbf{A self-evolution framework for physical coding agents.} We introduce EmbodiedEvo, a self-evolution coding agent. The agent autonomously observes, writes, and executes control programs, and retains exploration experience to improve subsequent physical task solving while model weights and the robot interface remain fixed.
    \item \textbf{Transferable physical experience from a single scene.} Skills preserve executable procedures, while Lessons retain operating preconditions, failure modes, and recovery guidance, capturing both how to act and when an operation applies. Experience acquired on one Playground scene per task is frozen for independent Test agents, enabling evaluation of its reuse across unseen layouts and different models.
    \item \textbf{A systematic analysis of the benefits and limits of evolution.} On RoboDojo-Sim, EmbodiedEvo achieves 44.48 Average Score and 37.82\% success. Further analyses show that unfinished exploration can yield useful knowledge and that weaker executors can use experience they struggle to acquire. Skills and Lessons retain comparable task performance to raw interaction history at 53.0\% lower Test execution cost, while precise contact control remains challenging.
\end{itemize}

\section{Related Work}
\label{sec:related}

\noindent\textbf{Foundation Models and Agentic Robot Control.}
Generalist policies acquire control capabilities through robot data and large-scale pretraining. RT-1, RT-2, and OpenVLA develop scalable visuomotor policies and transfer vision--language knowledge to actions~\citep{brohan2022rt1,brohan2023rt2,kim2024openvla}. The $\pi_0$ family extends this direction through flow-matching control, heterogeneous co-training, and learning from deployment experience~\citep{black2024pi0,physicalintelligence2025pi05,physicalintelligence2025pi06}. World-action models such as DreamZero and Cosmos Policy jointly model actions and future observations, connecting predictive dynamics with control~\citep{ye2026dreamzero,kim2026cosmospolicy}. These methods place substantial control knowledge in learned parameters.
A complementary line gives language models control over planning and execution. SayCan grounds skill selection in affordances, Inner Monologue incorporates execution feedback, and VoxPoser generates spatial value maps for motion planning~\citep{ahn2022saycan,huang2022innermonologue,huang2023voxposer}. HiRobot coordinates high-level reasoning with a low-level VLA~\citep{shi2025hirobot}. Direct control is explored by zero-shot trajectory generation, Agent as Policy, and the Astra evaluation~\citep{kwon2023trajectory,jia2026agentaspolicy,zhang2026unexpected}. Agent as Policy also reports faster repeated execution through saved procedures. EmbodiedEvo builds on this autonomy, studying how physical experience improves a coding agent operating through a fixed native interface without a learned motor policy.

\noindent\textbf{Programmatic Control and Reusable Skills.}
Code as Policies expresses robot behavior through executable programs, while Code as World represents physical structure and dynamics in executable form~\citep{liang2022codeaspolicies,wang2026codeasworlds}. ReAct interleaves reasoning with actions, and SWE-agent develops the workspace interface for iterative inspection, editing, and execution~\citep{yao2022react,yang2024sweagent}. In robotics, CaP-X studies how abstraction and execution feedback affect coding agents, EmbodiedSWE develops programs for long-horizon dexterous tasks, and RAPID turns visual demonstrations into relational robot programs~\citep{fu2026capx,shen2026embodiedswe,liu2026rapid}. HexaAnything combines explicit task state with perception, planning, and control tools~\citep{gao2026physicalcoding}.
Programs also provide a medium for accumulating experience. Voyager builds an executable skill library in Minecraft~\citep{wang2023voyager}; RATs acquires and freezes robot skills through self-directed play, including transfer to other coding agents~\citep{zhang2026playful}. ASPIRE accumulates validated repairs across task sequences, while RoboSkill combines code and textual guidance with tactile feedback~\citep{lu2026aspire,xie2026roboskill}. Archived, execution-improved software can also initialize new robot policies~\citep{kuroki2026closedloop}. The role of generated code matters: Eureka evolves reward programs for subsequent reinforcement learning~\citep{ma2023eureka}, whereas our Skills execute directly during interaction. EmbodiedEvo uses its frozen library to guide a live coding agent that continues to observe, revise programs, and recover during Test.

\noindent\textbf{Experience, Memory, and Harness Evolution.}
Self-evolution can adapt models, memory, tools, and agent architectures~\citep{gao2025selfevolving}. Prior interaction can influence control through several representations. ICRT conditions on sensorimotor demonstrations, RoboPrompt encodes action examples as text, and RoboDawn uses demonstrations to ground interface use and task strategies~\citep{fu2024icrt,yin2024roboprompt,guo2026robodawn}. RoboTTT compresses long interaction histories into fast weights updated at inference~\citep{jiang2026robottt}. External memory provides another route: Reflexion retains verbal feedback, ExpeL extracts reusable insights, and ReasoningBank distills strategies from both successful and failed experiences~\citep{shinn2023reflexion,zhao2023expel,ouyang2025reasoningbank}. Embodied-RAG organizes spatial and semantic memory for retrieval~\citep{xie2024embodiedrag}, while MemSkill and MemEvolve improve memory operations and architectures~\citep{zhang2026memskill,zhang2025memevolve}. Harness evolution changes the procedures surrounding the model. SkillOpt optimizes skill documents, SHAPER evolves skills and context-construction code, and EmbodiSkill distinguishes incorrect guidance from failures to follow valid guidance~\citep{yang2026skillopt,wang2026shaper,ju2026embodiskill,chi2026ai4ai}. Continual Harness revises agent scaffolding online, while RegenHarness governs revisions through explicit evidence and regression checks~\citep{karten2026continual,wang2026regenharness}. For physical manipulation, Harness VLA uses memory to compose a frozen VLA with analytic primitives, and Zetta evolves runtime critics and recovery skills around a frozen policy~\citep{zhang2026harnessvla,ding2026zetta}. EmbodiedRSI selects informative physical experiments for code--skill co-evolution and learns memory-management decisions; RoboHarn-Evo revises hierarchical task and action knowledge~\citep{song2026embodiedrsi,bao2026roboharn}. Our focus is the transferable content of experience: we acquire Skills and Lessons on one Playground scene, freeze them before independent Tests, and examine their value across scenes, models, and tasks. Comparisons with raw history and analysis of unsuccessful exploration further test what this retained experience contributes.

\begingroup
% Keep medium-sized method figures with text instead of on sparse float pages.
\renewcommand{\floatpagefraction}{0.8}

\begin{figure}[!htbp]
    \centering
    \includegraphics[width=\linewidth]{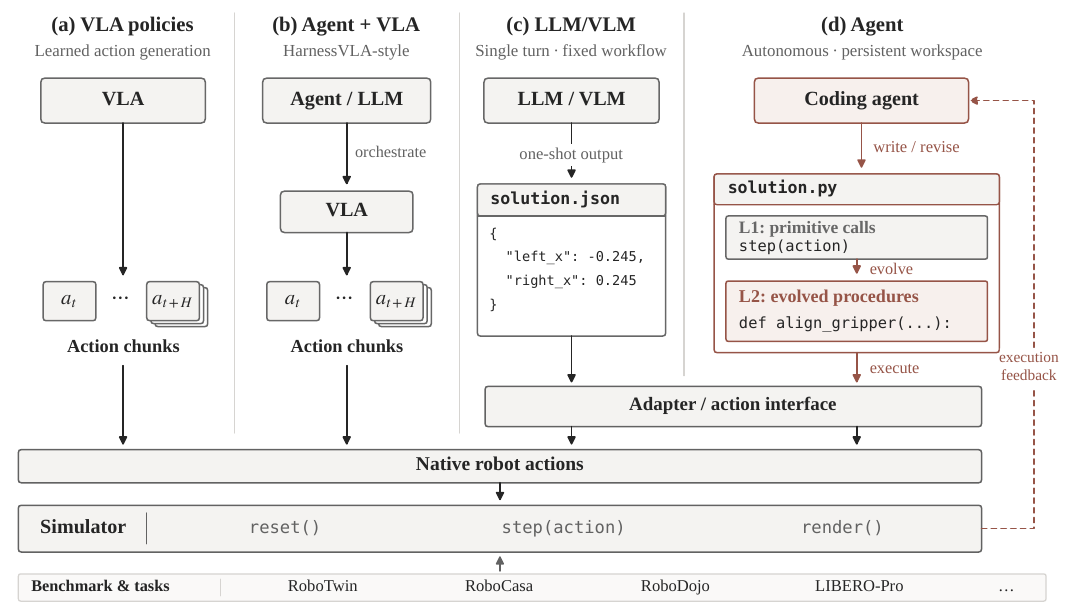}
    \caption{\textbf{Comparison of embodied control paradigms.} (a) A VLA policy emits native action chunks. (b) An agent orchestrates a VLA policy while retaining its action-generation role. (c) A single-turn LLM/VLM produces a structured action specification for an adapter. (d) A coding agent develops and revises executable programs through environmental feedback, composing primitive calls (L1) into evolved procedures (L2). The dashed loop shows execution feedback for program revision. }
    \label{fig:related-work-paradigms}
\end{figure}

\section{EmbodiedEvo}
\label{sec:method}

\subsection{Programmatic Robot Interface}
\label{sec:code}

The programmatic robot interface connects the coding agent to the physical environment through two levels of interaction. L1 provides the basic operations for observing and acting; L2 combines these operations into executable Python procedures. Figure~\ref{fig:related-work-paradigms} places this programmatic control within the broader set of embodied control paradigms.

\paragraph{L1: environment primitives.}
The environment exposes four primitives for reading the task, observing the scene, executing an action, and resetting the environment (Table~\ref{tab:primitives}). These operations form the fixed interface available to the agent. Reset is restricted to Playground, where it enables another attempt on the same scene.

\begin{table}[!htbp]
    \centering
    \small
    \setlength{\tabcolsep}{6pt}
    \renewcommand{\arraystretch}{1.15}
    \caption{\textbf{Native environment interface.} The first three primitives are available in both phases; reset is restricted to Playground phase.}
    \label{tab:primitives}
    \begin{tabularx}{\linewidth}{@{}p{0.25\linewidth}X@{}}
        \toprule
        \textbf{Primitive} & \textbf{Function} \\
        \midrule
        \texttt{get\_instruction()} & Returns the current task and goal. \\
        \texttt{get\_observation()} & Returns camera images and robot state without advancing physics. \\
        \texttt{step(action)} & Executes a native action and returns the resulting observation, completion signal, and termination status. \\
        \texttt{reset()} & Restores the same scene for another exploration attempt and renews its native action allowance. \\
        \bottomrule
    \end{tabularx}
\end{table}

In RoboDojo, observations contain RGB images and robot state, including joint states, end-effector poses, and gripper command states. Actions specify joint or end-effector targets together with gripper commands; the native controller handles inverse kinematics and interpolation. Object ground-truth poses, depth, and camera calibration are unavailable through this interface.

\paragraph{L2: programs composed from primitives.}
The agent writes Python programs that combine L1 calls through variables, functions, loops, and conditional branches. For example, a grasping procedure can read the current robot state, move towards a target, close the gripper, and lift. It can accept target coordinates as arguments and check the observed end-effector pose before continuing. A single program can therefore coordinate multiple native actions, while its functions can be reused in later programs.

This separation fixes what the robot can observe and execute while letting the agent decide how to compose those capabilities. Code makes the sequence, parameters, and checks explicit, so the agent can inspect and modify the procedure as the task progresses.
Figure~\ref{fig:code-to-task} shows how these programs connect to a task environment. The program-level entry point \texttt{step(code)} executes submitted Python through an adapter, and the program issues native \texttt{step(action)} calls to the selected simulator. The adapter keeps the interface consistent across task backends; our experiments use RoboDojo.

\FloatBarrier
\subsection{Closed-Loop Control with Coding Agents}
\label{sec:agency}

\begin{figure}[!htbp]
    \centering
    \includegraphics[width=0.94\linewidth]{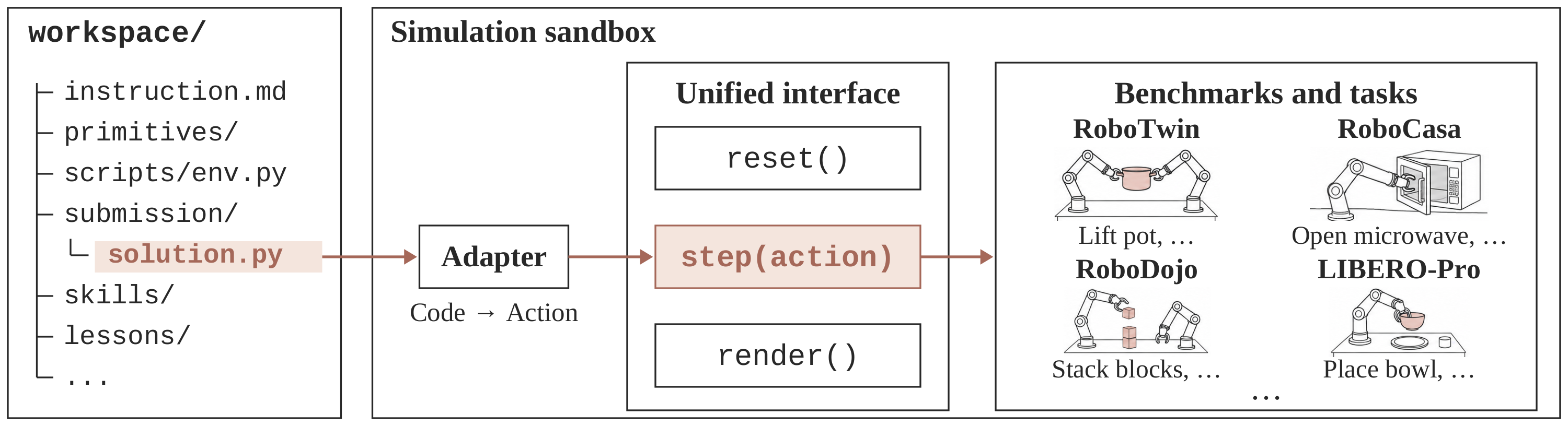}
    \caption{\textbf{Code-to-task mapping through a unified simulation interface.} The agent develops \texttt{submission/solution.py} within its workspace and submits it to an adapter. The adapter executes the program and issues native \texttt{step(action)} calls through the unified interface to the selected benchmark backend.}
    \label{fig:code-to-task}
\end{figure}

Agent as Policy assigns the control decisions to the coding agent: what to observe, which program to execute, and how to revise it after feedback. Using the L1/L2 interface, the agent directs an iterative process of inspection, programming, execution, and repair (Figure~\ref{fig:related-work-paradigms}d). It chooses the task decomposition and the timing of each step from the current observations and execution history.

\Needspace{4\baselineskip}
\paragraph{A workspace for acting through code.}
The agent works in an isolated filesystem workspace (Table~\ref{tab:workspace}). Task instructions and interface documentation provide the fixed specification; source files and scratch space support program development. Execution records preserve what ran and what happened, allowing the agent to inspect earlier attempts. Skills and Lessons provide the persistent knowledge introduced in Section~\ref{sec:harness}.

\begin{table}[!htbp]
    \centering
    \small
    \setlength{\tabcolsep}{6pt}
    \renewcommand{\arraystretch}{1.12}
    \caption{\textbf{Software-engineering workspace.} Paths are relative to the agent's workspace root. Protocol files and execution records are read-only to the agent.}
    \label{tab:workspace}
    \begin{tabularx}{\linewidth}{@{}p{0.37\linewidth}X@{}}
        \toprule
        \textbf{Component} & \textbf{Role and access} \\
        \midrule
        \texttt{instruction.md} & Fixed task delegation, phase rules, and budgets. \\
        \texttt{primitives/primitives.md} & Fixed documentation of the robot interface. \\
        \texttt{submission/solution.py} & Editable Python source for the next control segment. \\
        \texttt{scripts/env.py} & Fixed client for execution, status queries, and session completion. \\
        \texttt{observations/} & Host-written history of submitted source, observations, video, and execution logs. \\
        \texttt{scratch/} & Writable plans, calculations, and temporary analyses. \\
        \texttt{skills/}, \texttt{lessons/} & Reusable procedures and diagnostic knowledge; writable in Playground and read-only in Test. \\
        \bottomrule
    \end{tabularx}
\end{table}

\paragraph{An agent-directed interaction loop.}
The agent reads the task and inspects the scene, edits \texttt{submission/solution.py}, and submits it through the execution client (Figure~\ref{fig:code-to-task}). After execution, it can examine the submitted source, camera observations, standard output and errors, completion signal, and remaining budgets. It then decides whether to gather another observation, adjust a target, revise a function, continue to the next stage, or finish the episode. For example, after observing a dropped object, it can inspect the new scene and revise the grasp before attempting another lift. These choices are made within one coding-agent session.

\paragraph{Adapting to the current physical state.}
Robot state, Python variables, and function definitions persist between submissions. Editing code or encountering an error does not undo completed actions, so the agent must adapt its next program to the resulting scene. Playground permits same-scene resets and renews the native action allowance, while execution slots and elapsed time remain cumulative. Test disables reset, requiring recovery within the current scene and remaining budget. The workspace thus supports local program revision throughout an episode; Section~\ref{sec:harness} explains how experience can also be retained across episodes.

% \FloatBarrier
\subsection{Deep Dive into Astra's Failures}
\label{sec:motivation}

To understand where coding agents struggle in physical interaction, we evaluate GPT-6-Astra with xhigh reasoning effort in the code workspace described in Section~\ref{sec:agency}. The evaluation is on RoboDojo-sim tasks spanning Generalization, Precision, Long-Horizon, Memory, and Open, with 5 scenes per task. Astra receives the task description and robot-interface documentation, without the task wiki described in RoboProbe~\citep{roboprobe2026codebase}. Each episode starts with a fresh agent, workspace, and simulator, permits up to 100 accepted code executions within the task's native action allowance, and disables reset.

\begin{figure}[!htbp]
    \centering
    \includegraphics[width=0.90\linewidth]{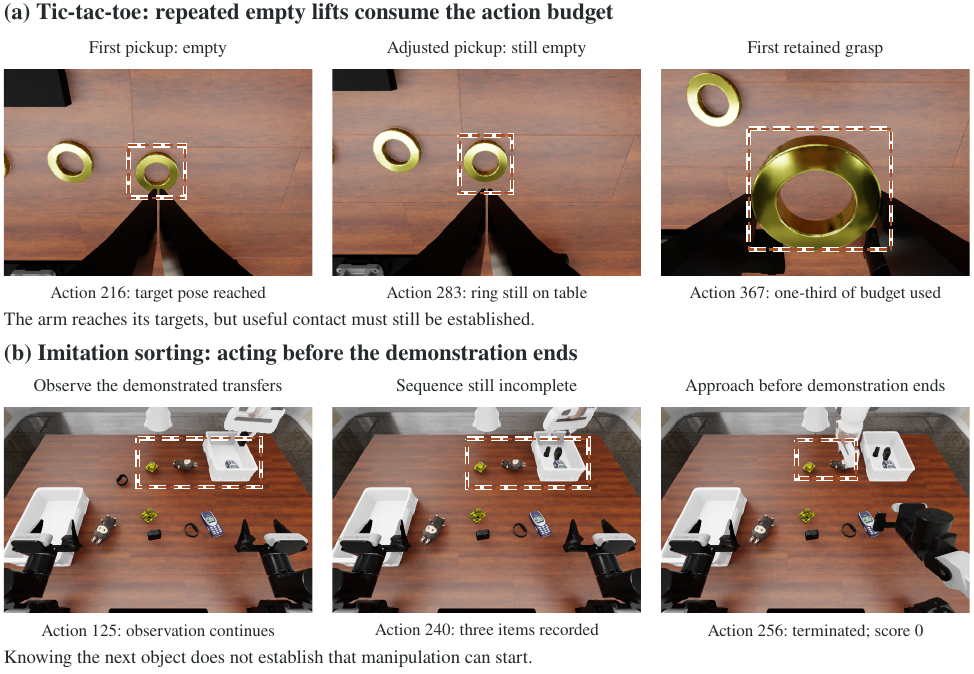}
    \par\vspace{6pt}
    % Astra task averages from the frozen summary.json.
\begingroup
\small
\setlength{\tabcolsep}{9pt}
\begin{tabular}{@{}lrr@{}}
\toprule
Task & Score & SR (\%) \\
\midrule
Tic-tac-toe & 55 & 0 \\
Imitation sorting & 10 & 0 \\
\bottomrule
\end{tabular}
\endgroup

    \caption{\textbf{Two failure trajectories of Astra.} (a) Two empty lifts precede the first retained grasp; the episode later exhausts its budget without completion. (b) Observation switches to manipulation during an unfinished demonstration, followed by termination. Boxes identify relevant objects. The table reports Astra's averages over all tested scenes of each task.}
    \label{fig:baseline-failures}
\end{figure}

Astra achieves an average success rate of 8.50\%. We examine its submitted programs, camera observations, and feedback to locate where plausible plans fail during physical execution. Figure~\ref{fig:baseline-failures} shows two illustrative episodes, each exposing a different gap between issuing a valid command and advancing the task.

\Needspace{4\baselineskip}
\paragraph{A correct action sequence still lacks a reliable manipulation routine.}
In tic-tac-toe, Astra attempts to pick up a ring and place it on the board. The gripper reaches its target positions, yet two successive attempts leave the ring on the table (actions 216 and 283). After inspecting a wrist-camera view and adjusting the grasp, Astra first visibly retains the ring at action 367, consuming one-third of its 1,100-action budget. The episode later times out without completing the board.
This trace shows how constructing a local manipulation routine can compete with task completion. It motivates retaining a parameterized procedure that combines approach, alignment, closure, and verification of object pickup, so that subsequent attempts can build on a working operation.

\Needspace{4\baselineskip}
\paragraph{Action decisions lack explicit precondition checks.}
In imitation sorting, Astra has identified only the first three objects at action 240, yet starts approaching the first \emph{while the demonstration continues}. The episode terminates 16 actions later, leaving 1,344 actions unused. 
The visible gap is acting without checking whether the demonstration has finished. Identifying a target alone does not establish readiness to manipulate it. This motivates retaining observation-grounded guidance on what to verify before acting, with unresolved explanations marked as hypotheses. Here, the trace supports checking demonstration completion; it does not prove that early movement caused termination or that waiting would ensure success.

The two failures expose complementary needs: (1) reliable procedures for carrying out an operation, and (2) explicit conditions for deciding when to attempt it. Within an episode, physical feedback can reveal both, but each fresh Astra agent starts again from the interface and task description. This motivates a mechanism that turns local corrections into reusable procedures and retains failure conditions to guide later decisions. Section~\ref{sec:harness} develops that mechanism through harness evolution.
\FloatBarrier

\subsection{Self-Improvement through Harness Evolution}
\label{sec:harness}
\label{sec:transfer}

EmbodiedEvo addresses the two needs in Section~\ref{sec:motivation} by retaining \emph{Skills} and \emph{Lessons} in an evolving harness (Figure~\ref{fig:method}). Skills combine fixed native primitives into reusable control procedures, such as grasping with a checked lift. Lessons record applicability conditions, observed failure modes, and guidance for diagnosis and recovery. Together, they capture how to carry out an operation and what to check before or after it. Self-improvement updates these artifacts while model parameters, task goals, and the robot interface remain fixed.

\begin{figure}[!htbp]
    \centering
    \includegraphics[width=0.9\linewidth]{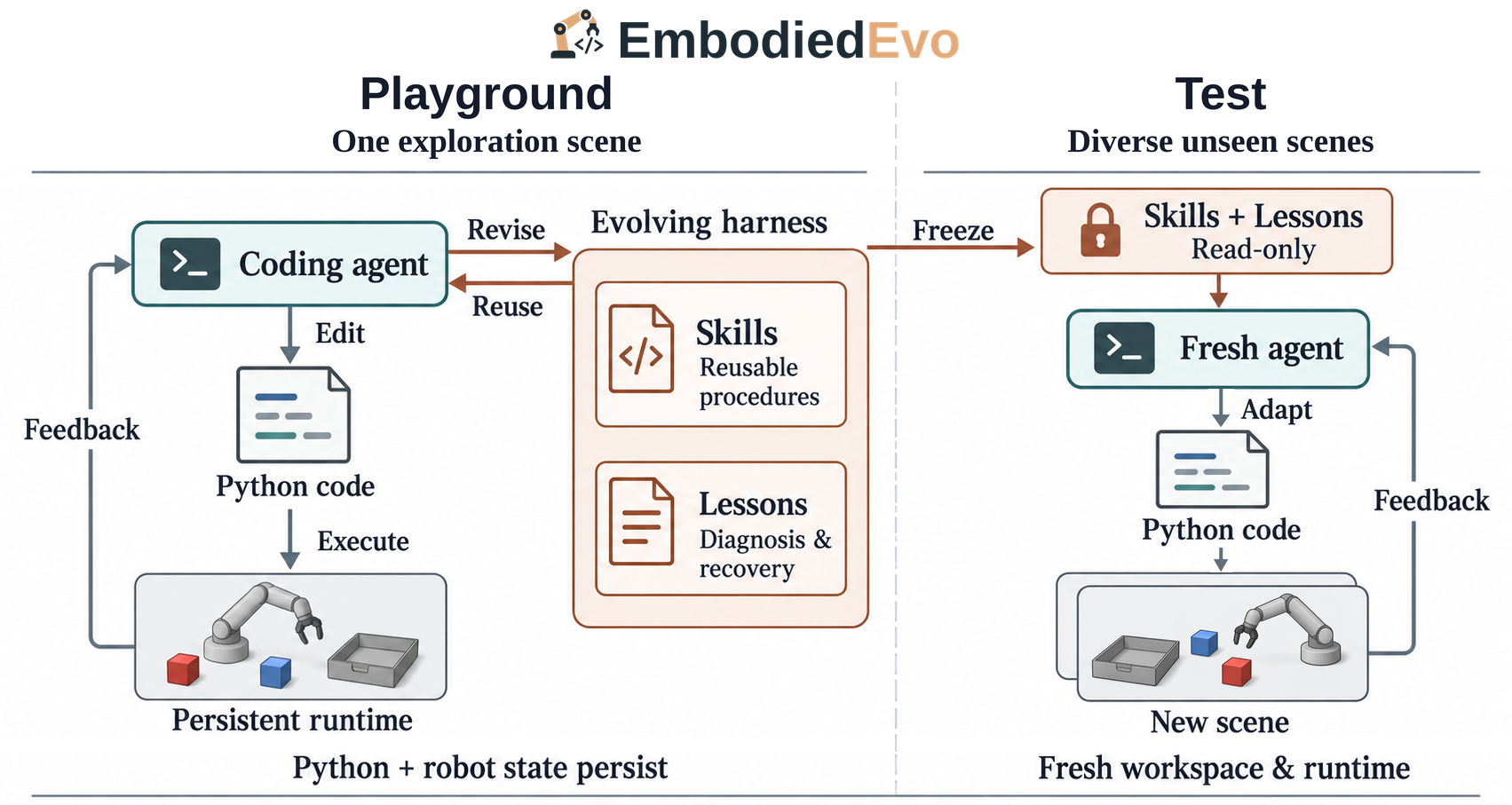}
    \caption{\textbf{EmbodiedEvo framework.} Left: on one Playground scene, the agent edits and executes code, uses feedback to revise Skills and Lessons, and reuses them in subsequent attempts. The Freeze arrow transfers these artifacts to Test. Right: each unseen scene starts with a fresh agent and runtime; code adapts through feedback while the harness stays read-only. Robot scenes are schematic.}
    \label{fig:method}
\end{figure}

On the left of Figure~\ref{fig:method}, one Playground scene hosts a persistent loop of code editing, execution, and feedback. Robot state and Python definitions persist between submissions; reset allows another attempt on the same scene. The Revise and Reuse arrows show how the agent updates Skills and Lessons and applies them in subsequent attempts. A successful operation can contribute a parameterized procedure; an unsuccessful attempt can reveal a useful condition or suggest a recovery strategy. Even an unfinished task can yield useful findings; tentative explanations remain hypotheses.

\Needspace{2\baselineskip}
The Freeze arrow marks the end of Playground: only Skills and Lessons are carried into Test as read-only artifacts. On the right, each unseen scene of the same task starts with a fresh agent, workspace, Python namespace, and simulator. The agent uses the frozen harness to construct code and adapts that code through execution feedback. Reset is disabled, and Test episodes do not update the harness or share new experience. This separates within-scene adaptation from the contribution of knowledge acquired during Playground.

The same frozen artifacts can be supplied to another executor because their contents are explicit procedures and guidance. Cross-model reuse requires no additional Playground exploration, parameter updates, or learned adapter. We compare self-generated and transferred harnesses under the same executor and Test conditions to assess whether the acquired knowledge remains useful across models.
\FloatBarrier

\endgroup
\begingroup
\compactfloatlayout
\newtcolorbox{researchquestions}{
    enhanced,
    colback=brandbg,
    colframe=brandprimary,
    colbacktitle=brandprimary,
    coltitle=brandfg,
    fonttitle=\bfseries,
    title={Research questions},
    arc=1mm, boxrule=0.5pt,
    left=8pt, right=8pt, top=4pt, bottom=4pt,
    toptitle=2pt, bottomtitle=2pt,
    before skip=6pt, after skip=6pt
}

\section{Experiments}
\label{sec:experiments}

We evaluate EmbodiedEvo on RoboDojo-Sim to address the following research questions: \textbf{(1)} How well does EmbodiedEvo perform compared with VLA- and WAM-based methods? \textbf{(2)} Which capabilities improve through harness evolution, and which tasks remain difficult? \textbf{(3)} Can acquired experience transfer across models and tasks?
% \textbf{(4)} How do Skills and Lessons compare with raw interaction history as representations of experience? \textbf{(5)} How do experience and its representation affect execution expense?

\subsection{Experimental Setup}
\label{sec:setup}

\paragraph{Tasks and agents.}
RoboDojo-Sim~\citep{chen2026robodojo} is a challenging benchmark for evaluating generalist robot manipulation policies. It comprises 12 Generalization, 8 Precision, 8 Long-Horizon, 6 Memory, and 8 Open tasks. Each harness(skills + lessons) is acquired on one Playground scene layout distinct from the Test layouts. Executors are GPT-6-Astra at xhigh and medium effort, GPT-5.6-sol at xhigh, and GPT-5.6-luna at xhigh, abbreviated Astra, Sol, and Luna. 

\paragraph{Evaluation scales.}
We use two evaluation scales: (1) dev set: 5 scenes per task (210 episodes per condition) for input comparisons, ablations, and cross-model analyses; and (2) full set: 50 scenes per task (2,100 episodes, one seed) for Full Evaluation of the complete benchmark. Paired comparisons use identical scenes. For each Generalization task, the two scales use three standard and two random scenes, and 25 standard and 25 random scenes, respectively. Scene identities follow the public evaluation records~\citep{roboprobe2026reference}. Cross-task transfer uses a fixed subset of 12 target tasks and is reported separately from the full-benchmark averages.

\paragraph{Inputs and budgets.}
The task description states the goal in one sentence; the official wiki provides additional task rules and operational guidance~\citep{roboprobe2026recipes}. Full Evaluation uses \emph{task description + wiki} to align task information with the official GPT-6-Astra evaluation~\citep{zhang2026unexpected}. The input ablation compares both settings; all other paired analyses use \emph{pure task description}.
Each Playground chain and Test episode allows up to 100 accepted code executions and retains the task's native action budget. Playground permits same-scene resets; acquired Skills and Lessons are then frozen. Test starts a fresh agent, workspace, and simulator per scene, with read-only experience and no reset. 

\paragraph{Metrics.}
We report Score (0--100) and success rate (SR, \%). Score quantifies task completion, including partial progress, according to the benchmark's task-specific scoring rules. SR is the percentage of episodes that satisfy the task's success criterion.

\subsection{Overall Performance}
\label{sec:performance}
% Keep the main result table below its subsection heading on this page.
\suppressfloats[t]

% \begin{researchquestions}
% Can a self-improving coding agent compete with VLA- and WAM-based systems across diverse manipulation tasks?
% \end{researchquestions}

We evaluate EmbodiedEvo with Astra/xhigh on the full RoboDojo-Sim set. We compare category-balanced Score and SR with the VLA, WAM, and agent-based results in the October 9, 2026 leaderboard snapshot.

Table~\ref{tab:full-results} shows that EmbodiedEvo achieves the highest Average Score (44.48) and SR (37.82\%) among the listed methods. It exceeds VPP2, the strongest listed WAM in aggregate, by 5.22 Score points and 5.56 SR percentage points. Against Simate-beta (VLA), the margins are 10.53 and 9.86 points, respectively; it also exceeds PhysicalRSI (Agent + VLA) by 8.21 and 6.44 points.

% Source: https://robodojo-benchmark.com/leaderboard (updated 2026-10-09; accessed 2026-10-12).
% Ours: runs/final/202610061832_robodojo_D_recipe_astra_xhigh_v01_a01/summary.md in EAHarness.
% All rows ascend by Average Score; Ours reports the completed seed-1 evaluation.
% Layout references: OpenVLA, Table 1 (https://arxiv.org/abs/2406.09246),
% and Octo, Table 1 (https://octo-models.github.io/paper.pdf).
\begin{table}[!htbp]
\centering
\begingroup
% Shared typography and palette, matching the full-benchmark main table.
% Input inside a local group so the paper's body typography is unchanged.
\renewcommand{\rmdefault}{ptm}
\renewcommand{\sfdefault}{ptm}
\rmfamily
\captionsetup{font=small,skip=6pt}
\definecolor{lbblue}{HTML}{315F85}
\definecolor{lbteal}{HTML}{39746C}
\definecolor{lbviolet}{HTML}{6F598A}
\definecolor{lborange}{HTML}{9A683D}
\definecolor{lbmuted}{HTML}{59616A}
\definecolor{lbaverage}{HTML}{F1F5FA}
\definecolor{lbours}{HTML}{FAF4EE}
\newcommand{\lbbest}[1]{{\bfseries #1}}
\newcommand{\lbtype}[2]{\textcolor{#1}{\mbox{\fontsize{7.2}{9}\selectfont #2}}}
\newcommand{\lbhead}[1]{\makebox[50pt][c]{\bfseries #1}}
\newcommand{\lbpair}[2]{\makebox[23pt][r]{#1}\hspace{4pt}\makebox[23pt][r]{#2}}
\newcommand{\lbmetrics}{\lbpair{\textcolor{lbmuted}{\fontsize{7}{8}\selectfont Score}}{\textcolor{lbmuted}{\fontsize{7}{8}\selectfont SR (\%)}}}
\fontsize{8}{10}\selectfont
\setlength{\tabcolsep}{3.4pt}
\renewcommand{\arraystretch}{1.6}
\setlength{\heavyrulewidth}{0.65pt}
\setlength{\lightrulewidth}{0.35pt}
\setlength{\aboverulesep}{3pt}
\setlength{\belowrulesep}{3pt}

\caption{\textbf{Full RoboDojo simulation evaluation.} EmbodiedEvo alongside selected systems from the October 9, 2026 leaderboard snapshot, ordered by increasing Average Score. Score and success rate (SR, \%) are higher-is-better; bold marks the best value in each metric column. }
\label{tab:full-results}
\begin{tabularx}{\linewidth}{>{\raggedright\arraybackslash}X l >{\columncolor{lbaverage}}c *{5}{c}}
\toprule
\multirow{2}{*}{\bfseries Model} & \multirow{2}{*}{\bfseries Type} & \textcolor{lbblue}{\lbhead{Average}} & \lbhead{Generalization} & \lbhead{Precision} & \lbhead{Long-horizon} & \lbhead{Memory} & \lbhead{Open} \\
& & \lbmetrics & \lbmetrics & \lbmetrics & \lbmetrics & \lbmetrics & \lbmetrics \\
\midrule
% RoboDojo Team
GPT-6-Astra & \lbtype{lbblue}{Agent} & \lbpair{28.97}{22.48} & \lbpair{33.36}{\lbbest{30.50}} & \lbpair{12.65}{4.00} & \lbpair{21.45}{8.25} & \lbpair{43.04}{38.67} & \lbpair{34.36}{31.00} \\
% Shanghai AI Lab
InternW0-$\Delta$ & \lbtype{lbteal}{WAM} & \lbpair{30.77}{23.91} & \lbpair{30.09}{22.78} & \lbpair{31.98}{23.25} & \lbpair{46.29}{29.33} & \lbpair{34.67}{34.00} & \lbpair{10.84}{10.17} \\
% Simate
Simate-beta & \lbtype{lbviolet}{VLA} & \lbpair{33.95}{27.96} & \lbpair{35.09}{27.95} & \lbpair{34.35}{26.92} & \lbpair{\lbbest{57.84}}{\lbbest{43.42}} & \lbpair{33.33}{33.00} & \lbpair{9.12}{8.50} \\
% Awomo
Awomo-0.5 & \lbtype{lbteal}{WAM} & \lbpair{35.34}{29.64} & \lbpair{30.31}{23.84} & \lbpair{39.65}{33.50} & \lbpair{40.36}{26.75} & \lbpair{42.27}{41.22} & \lbpair{24.09}{22.92} \\
% HKU MMLab & KAI
PhysicalRSI & \lbtype{lborange}{Agent + VLA} & \lbpair{36.27}{31.38} & \lbpair{21.30}{15.62} & \lbpair{38.05}{32.50} & \lbpair{46.37}{37.33} & \lbpair{46.74}{46.56} & \lbpair{28.88}{24.92} \\
% RobotEra
VPP2 & \lbtype{lbteal}{WAM} & \lbpair{39.26}{32.26} & \lbpair{\lbbest{36.38}}{29.50} & \lbpair{\lbbest{43.03}}{\lbbest{35.17}} & \lbpair{56.28}{37.00} & \lbpair{56.01}{55.56} & \lbpair{4.61}{4.08} \\
\midrule
\rowcolor{lbours}
{\bfseries EmbodiedEvo (Ours)} & \lbtype{lbblue}{Agent} & \lbpair{\lbbest{44.48}}{\lbbest{37.82}} & \lbpair{34.12}{29.67} & \lbpair{23.65}{13.25} & \lbpair{37.20}{23.50} & \lbpair{\lbbest{71.90}}{\lbbest{69.67}} & \lbpair{\lbbest{55.52}}{\lbbest{53.00}} \\
\bottomrule
\end{tabularx}
\par\vspace{3pt}
\endgroup
\end{table}

\textbf{EmbodiedEvo leads on Memory and Open tasks by large margins.} EmbodiedEvo achieves 71.90 Score / 69.67\% SR on Memory and 55.52 / 53.00\% on Open. Its SR exceeds the next-best listed method in each category: VPP2 at 55.56\% on Memory and GPT-6-Astra at 31.00\% on Open. The cases in Section~\ref{sec:capability} suggest why Skills and Lessons suit these tasks. Skills provide reusable grasping, carrying, and placement procedures that can be composed around the current instruction; Lessons guide observation, sequence tracking, and checks before advancing to the next phase. Together, they connect remembering what the current scene requires with procedures for executing it, supporting memory-dependent sequences and instruction-conditioned manipulation.

\textbf{Agent still struggles with precise manipulation and long-horizon execution.} Its Precision SR is 13.25\%, below VPP2's 35.17\%, while its Long-Horizon SR is 23.50\%, below Simate-beta's 43.42\%. These tasks require accurate contact and reliable execution across successive operations. The failure cases in Section~\ref{sec:capability} show that reaching a commanded pose can leave a grasp unsecured, and that completing some stages can leave the overall task unfinished. Retained procedures and guidance therefore remain dependent on reliable physical execution: contact accuracy and maintaining successful manipulation across a sequence are continuing bottlenecks.

\FloatBarrier
\subsection{What Does Harness Evolution Actually Change?}
\label{sec:capability}

\begin{researchquestions}
\textbf{(1)} What can a coding agent already solve without evolution? \textbf{(2)} What new capabilities does evolution add? \textbf{(3)} What remains unsolved after evolution?
\end{researchquestions}

To answer the questions above, we conduct paired evaluations of Astra/xhigh with and without evolved experience, using identical scenes and task inputs. We test both wiki settings~\citep{roboprobe2026recipes}, then examine task outcomes and execution traces to distinguish existing competence, newly acquired capabilities, and remaining failures. 
Due to the cost of large-scale agent evaluation, we conduct this and all subsequent experiments on the \emph{dev set (5 scenes per task)}.

\textbf{EmbodiedEvo consistently improves performance across different input settings.} In Table~\ref{tab:task-knowledge}, EmbodiedEvo improves Average Score by 23.87 points and SR by 21.67 percentage points without wiki. With wiki supplied to both methods, the gains remain 21.75 Score points and 21.72 SR points. Figure~\ref{fig:task-knowledge} shows positive Score and SR gains across all five categories under both inputs. These gains suggest that experience contributes execution knowledge under either input setting; the following traces show how Skills supply tested operations and Lessons guide their use.

% Generated from audited B/D and B_recipe/D_recipe outcomes.
\begin{table}[!htbp]
\centering\begingroup
% Shared typography and palette, matching the full-benchmark main table.
% Input inside a local group so the paper's body typography is unchanged.
\renewcommand{\rmdefault}{ptm}
\renewcommand{\sfdefault}{ptm}
\rmfamily
\captionsetup{font=small,skip=6pt}
\definecolor{lbblue}{HTML}{315F85}
\definecolor{lbteal}{HTML}{39746C}
\definecolor{lbviolet}{HTML}{6F598A}
\definecolor{lborange}{HTML}{9A683D}
\definecolor{lbmuted}{HTML}{59616A}
\definecolor{lbaverage}{HTML}{F1F5FA}
\definecolor{lbours}{HTML}{FAF4EE}
\newcommand{\lbbest}[1]{{\bfseries #1}}
\newcommand{\lbtype}[2]{\textcolor{#1}{\mbox{\fontsize{7.2}{9}\selectfont #2}}}
\newcommand{\lbhead}[1]{\makebox[50pt][c]{\bfseries #1}}
\newcommand{\lbpair}[2]{\makebox[23pt][r]{#1}\hspace{4pt}\makebox[23pt][r]{#2}}
\newcommand{\lbmetrics}{\lbpair{\textcolor{lbmuted}{\fontsize{7}{8}\selectfont Score}}{\textcolor{lbmuted}{\fontsize{7}{8}\selectfont SR (\%)}}}
\fontsize{8}{10}\selectfont
\setlength{\tabcolsep}{3.4pt}
\renewcommand{\arraystretch}{1.6}
\setlength{\heavyrulewidth}{0.65pt}
\setlength{\lightrulewidth}{0.35pt}
\setlength{\aboverulesep}{3pt}
\setlength{\belowrulesep}{3pt}

\caption{\textbf{Task knowledge and inherited experience.} All rows use Astra/xhigh on matched scenes. Each cell reports Score / SR (\%).}
\label{tab:task-knowledge}
\setlength{\tabcolsep}{2pt}
\renewcommand{\arraystretch}{1.4}
\renewcommand{\lbhead}[1]{\makebox[44pt][c]{\bfseries #1}}
\renewcommand{\lbpair}[2]{\makebox[21pt][r]{#1}\hspace{2pt}\makebox[21pt][r]{#2}}
\begin{tabularx}{\linewidth}{l >{\raggedright\arraybackslash}X >{\columncolor{lbaverage}}c *{5}{c}}
\toprule
\multirow{2}{*}{\bfseries Method} & \multirow{2}{*}{\bfseries Input} & \textcolor{lbblue}{\lbhead{Average}} & \lbhead{Generalization} & \lbhead{Memory} & \lbhead{Precision} & \lbhead{Long-horizon} & \lbhead{Open} \\
& & \lbmetrics & \lbmetrics & \lbmetrics & \lbmetrics & \lbmetrics & \lbmetrics \\
\midrule
Astra & Task description & \cellcolor{lbaverage}\lbpair{14.77}{8.50} & \lbpair{3.19}{0.00} & \lbpair{12.67}{10.00} & \lbpair{11.38}{5.00} & \lbpair{18.12}{0.00} & \lbpair{28.50}{27.50} \\
\rowcolor{lbours}
EmbodiedEvo (Astra) & Task description & \cellcolor{lbaverage}\lbpair{\lbbest{38.64}}{\lbbest{30.17}} & \lbpair{22.36}{16.67} & \lbpair{59.33}{56.67} & \lbpair{29.88}{15.00} & \lbpair{29.50}{12.50} & \lbpair{52.12}{50.00} \\
\midrule
Astra & Task description + wiki & \cellcolor{lbaverage}\lbpair{20.67}{13.36} & \lbpair{12.33}{7.64} & \lbpair{39.67}{36.67} & \lbpair{15.00}{7.50} & \lbpair{19.38}{0.00} & \lbpair{17.00}{15.00} \\
\rowcolor{lbours}
EmbodiedEvo (Astra) & Task description + wiki & \cellcolor{lbaverage}\lbpair{\lbbest{42.42}}{\lbbest{35.08}} & \lbpair{36.15}{31.25} & \lbpair{70.33}{66.67} & \lbpair{23.12}{12.50} & \lbpair{33.75}{20.00} & \lbpair{48.75}{45.00} \\
\bottomrule
\end{tabularx}
\par\vspace{3pt}\endgroup
\end{table}

\begin{figure}[!htbp]
\centering
\includegraphics[width=0.8\linewidth]{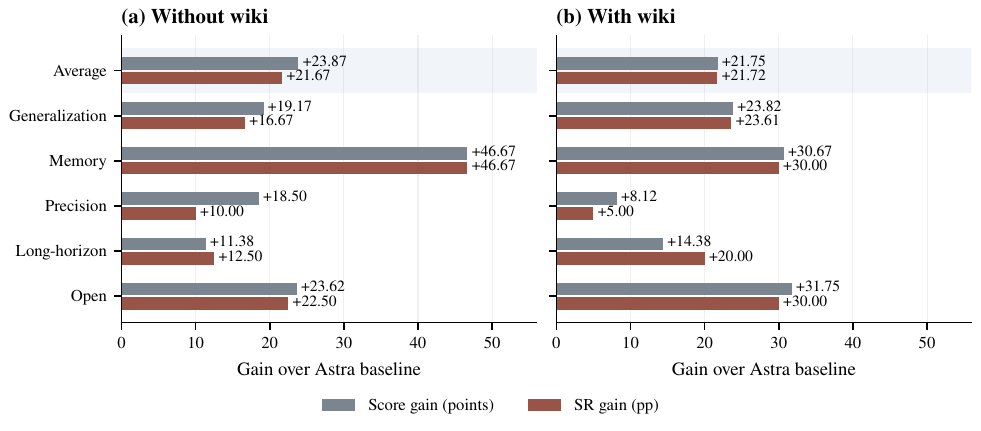}
\caption{\textbf{Effect of EmbodiedEvo with and without wiki.} Score (points) and SR (percentage points) gains over Astra/xhigh on matched scenes and inputs: (a) task description only; (b) task description + wiki.}
\label{fig:task-knowledge}
\end{figure}

We next examine the operations and decisions behind these gains. Table~\ref{tab:capability-results} organizes representative tasks into existing competence, newly acquired capability, and remaining limits. Appendix~\ref{app:task_catalog} provides the complete dev- and full-set results.

% Generated by ccfa-workfiles/experiments/current-results/write_capability_table.py.
\begin{table}[!htbp]
\centering\begingroup
% Shared typography and palette, matching the full-benchmark main table.
% Input inside a local group so the paper's body typography is unchanged.
\renewcommand{\rmdefault}{ptm}
\renewcommand{\sfdefault}{ptm}
\rmfamily
\captionsetup{font=small,skip=6pt}
\definecolor{lbblue}{HTML}{315F85}
\definecolor{lbteal}{HTML}{39746C}
\definecolor{lbviolet}{HTML}{6F598A}
\definecolor{lborange}{HTML}{9A683D}
\definecolor{lbmuted}{HTML}{59616A}
\definecolor{lbaverage}{HTML}{F1F5FA}
\definecolor{lbours}{HTML}{FAF4EE}
\newcommand{\lbbest}[1]{{\bfseries #1}}
\newcommand{\lbtype}[2]{\textcolor{#1}{\mbox{\fontsize{7.2}{9}\selectfont #2}}}
\newcommand{\lbhead}[1]{\makebox[50pt][c]{\bfseries #1}}
\newcommand{\lbpair}[2]{\makebox[23pt][r]{#1}\hspace{4pt}\makebox[23pt][r]{#2}}
\newcommand{\lbmetrics}{\lbpair{\textcolor{lbmuted}{\fontsize{7}{8}\selectfont Score}}{\textcolor{lbmuted}{\fontsize{7}{8}\selectfont SR (\%)}}}
\fontsize{8}{10}\selectfont
\setlength{\tabcolsep}{3.4pt}
\renewcommand{\arraystretch}{1.6}
\setlength{\heavyrulewidth}{0.65pt}
\setlength{\lightrulewidth}{0.35pt}
\setlength{\aboverulesep}{3pt}
\setlength{\belowrulesep}{3pt}

\small
\setlength{\tabcolsep}{4pt}
\renewcommand{\arraystretch}{1.1}
\caption{\textbf{Task performance before and after harness evolution.} Selected tasks with the same Astra/xhigh executor and matched dev scenes. Bold marks the higher value within each before/after pair; $\Delta$SR is the change in percentage points. Groups describe observed success in these scenes.}
\label{tab:capability-results}
\begin{tabularx}{\linewidth}{@{}>{\raggedright\arraybackslash}X l rr >{\columncolor{lbours}}r >{\columncolor{lbours}}r r@{}}
\toprule
\multirow{2}{*}{\bfseries Task} & \multirow{2}{*}{\bfseries Category} & \multicolumn{2}{c}{\bfseries Astra} & \multicolumn{2}{c}{\bfseries EmbodiedEvo (Astra)} & \multirow{2}{*}{$\Delta$SR (pp)} \\
\cmidrule(lr){3-4}\cmidrule(lr){5-6}
& & Score & SR (\%) & Score & SR (\%) & \\
\midrule
\multicolumn{7}{l}{\textcolor{lbmuted}{\bfseries (1) Successful before evolution}} \\
Align blocks & Open & 60 & 60 & \textbf{80} & \textbf{80} & $+20$ \\
Conveyor matching & Memory & 60 & 60 & \textbf{100} & \textbf{100} & $+40$ \\
\addlinespace[5pt]
\multicolumn{7}{l}{\textcolor{lbmuted}{\bfseries (2) Successful only after evolution}} \\
Stack blocks & Generalization & 10 & 0 & \textbf{100} & \textbf{100} & $+100$ \\
Tic-tac-toe & Long-Horizon & 55 & 0 & \textbf{100} & \textbf{100} & $+100$ \\
Imitate sorting & Memory & 10 & 0 & \textbf{90} & \textbf{80} & $+80$ \\
\addlinespace[5pt]
\multicolumn{7}{l}{\textcolor{lbmuted}{\bfseries (3) Unsuccessful in both conditions}} \\
Insert tubes & Precision & 4 & 0 & \textbf{36} & 0 & 0 \\
Insert key & Precision & 9 & 0 & \textbf{15} & 0 & 0 \\
\bottomrule
\end{tabularx}
\endgroup
\end{table}

\textbf{Visible goals and locally correctable actions can already be solved without evolution.} Astra achieves 60\% SR on both block alignment and conveyor matching; evolution raises these to 80\% and 100\%, respectively. Both tasks provide visible targets and feedback for adjusting the next action, allowing online coding and observation to support local corrections. Here, evolution improves the reliability of operations Astra can already perform.

\textbf{Evolution supplies reusable operations and conditions for advancing through a task.} Tic-tac-toe and block stacking rise from zero to 100\% SR; imitation sorting rises from zero to 80\%. These tasks require more than identifying a spatial goal: successful local operations must be repeated, or action must wait until the necessary scene information has been collected.

Figure~\ref{fig:capability-evolution} contrasts two such decision points on matched scenes. In tic-tac-toe, the baseline repeatedly lifts without retaining a ring, eventually scoring 50 with no success. EmbodiedEvo executes the inherited \texttt{rim\_pick} routine, retains a ring at action 38, and completes the board with Score 100. The useful change is an executable grasp procedure that supports repeated placements. In imitation sorting, the baseline starts moving before the demonstration finishes and terminates with Score 0. The evolved agent follows its observation lesson, records the current sequence through action 455, and reproduces it successfully. The order differs from Playground, so the reusable knowledge concerns \emph{when to act}, rather than a memorized object sequence. Together, these cases connect Skills to repeatable physical operations and Lessons to the conditions for using them. They illustrate the observed mechanisms; the paired runs do not isolate one artifact's causal effect. Appendix~\ref{app:capability-cases} provides the supporting records, and Appendix~\ref{app:agent-trace} gives a complete successful language-stacking trace.

\begin{figure}[!htbp]
\centering
\includegraphics[width=\linewidth]{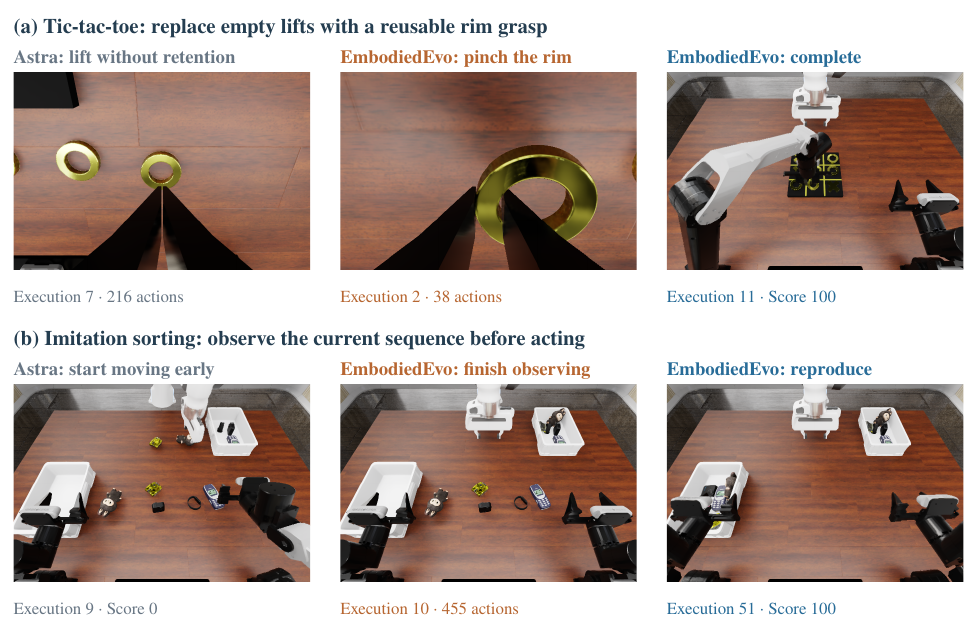}
\caption{\textbf{New capability: changes at the decisive operation or phase.} Each row compares Astra and EmbodiedEvo on the same task and scene. (a) Tic-tac-toe, layout 27: empty pickup versus an inherited rim grasp and completed board. The first pickups use different rings and arms. (b) Imitation sorting, layout 4: early motion versus observing the complete new order before executing it. Both timing and motion differ; the baseline feedback does not identify the termination cause. Frames show selected decision points, not synchronized times.}
\label{fig:capability-evolution}
\end{figure}

\Needspace{4\baselineskip}
\textbf{Reusable motion does not guarantee precise contact.} Insert tubes improves from 4 to 36 Score and insert key from 9 to 15, yet neither achieves a dev-set success. Figure~\ref{fig:capability-limit} localizes the key failure: bounded motion retains the key during pickup, but the receiving hand closes empty during handover. Subsequent corrections do not secure it; releasing the donor drops the key at the action limit, \emph{before insertion}. A working pickup procedure therefore leaves the next contact transition unresolved. This gap between successful subroutines and complete manipulation explains why partial Score gains need not become SR gains on Precision tasks.

\begin{figure}[!htbp]
\centering
\includegraphics[width=\linewidth]{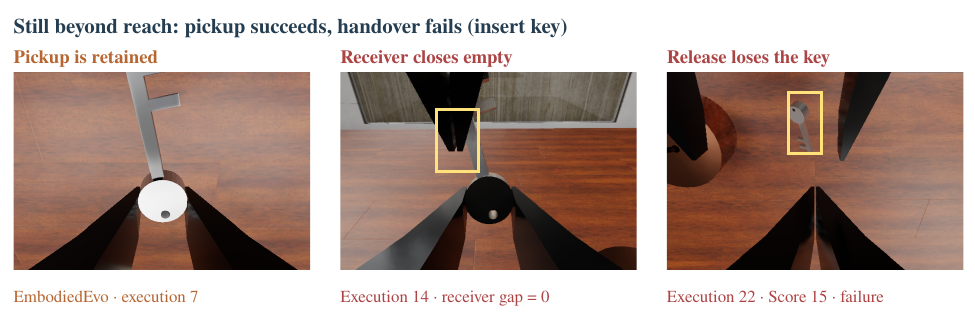}
\caption{\textbf{Remaining limit: unsuccessful contact transfer.} EmbodiedEvo, insert key, layout 44. The donor retains the key, but the receiver closes without grasping the shaft (highlighted); the final release loses the key (highlighted). Wrist observations and gripper feedback locate the failure at handover, before insertion. This episode scores 15 with no success, matching the task's dev-set mean.}
\label{fig:capability-limit}
\end{figure}

\subsection{Transferring Evolved Experience across Models and Tasks}
\label{sec:experience-transfer}

\begin{researchquestions}
\textbf{(1)} Can experience acquired by a stronger model improve weaker executors? \textbf{(2)} Can experience from other tasks improve performance on the current task?
\end{researchquestions}

To answer the questions above, we conduct matched comparisons of No Experience, self-evolved harnesses, and Astra/xhigh harnesses for each receiving executor. We also fix Astra/xhigh on twelve target tasks and compare same-task, related-task, and mismatched-task harnesses against No Experience. All transferred Skills and Lessons are frozen, with no further exploration by the receiver.

% Generated by ccfa-workfiles/experiments/current-results/write_tables.py.
\begin{table}[!htbp]
\centering
\begingroup
% Shared typography and palette, matching the full-benchmark main table.
% Input inside a local group so the paper's body typography is unchanged.
\renewcommand{\rmdefault}{ptm}
\renewcommand{\sfdefault}{ptm}
\rmfamily
\captionsetup{font=small,skip=6pt}
\definecolor{lbblue}{HTML}{315F85}
\definecolor{lbteal}{HTML}{39746C}
\definecolor{lbviolet}{HTML}{6F598A}
\definecolor{lborange}{HTML}{9A683D}
\definecolor{lbmuted}{HTML}{59616A}
\definecolor{lbaverage}{HTML}{F1F5FA}
\definecolor{lbours}{HTML}{FAF4EE}
\newcommand{\lbbest}[1]{{\bfseries #1}}
\newcommand{\lbtype}[2]{\textcolor{#1}{\mbox{\fontsize{7.2}{9}\selectfont #2}}}
\newcommand{\lbhead}[1]{\makebox[50pt][c]{\bfseries #1}}
\newcommand{\lbpair}[2]{\makebox[23pt][r]{#1}\hspace{4pt}\makebox[23pt][r]{#2}}
\newcommand{\lbmetrics}{\lbpair{\textcolor{lbmuted}{\fontsize{7}{8}\selectfont Score}}{\textcolor{lbmuted}{\fontsize{7}{8}\selectfont SR (\%)}}}
\fontsize{8}{10}\selectfont
\setlength{\tabcolsep}{3.4pt}
\renewcommand{\arraystretch}{1.6}
\setlength{\heavyrulewidth}{0.65pt}
\setlength{\lightrulewidth}{0.35pt}
\setlength{\aboverulesep}{3pt}
\setlength{\belowrulesep}{3pt}

\caption{\textbf{Performance under three experience conditions.} Score and success rate (SR, \%) are higher-is-better. Each executor uses matched scenes across conditions. Self-evolved harnesses are acquired by the listed executor; the final group reuses frozen experience from Astra/xhigh.}
\label{tab:paired-results}
\setlength{\tabcolsep}{2.5pt}
\begin{tabularx}{\linewidth}{>{\raggedright\arraybackslash}X >{\columncolor{lbaverage}}c *{5}{c}}
\toprule
\multirow{2}{*}{\bfseries Executor} & \textcolor{lbblue}{\lbhead{Average}} & \lbhead{Generalization} & \lbhead{Precision} & \lbhead{Long-horizon} & \lbhead{Memory} & \lbhead{Open} \\
& \lbmetrics & \lbmetrics & \lbmetrics & \lbmetrics & \lbmetrics & \lbmetrics \\
\midrule
\multicolumn{7}{l}{\textcolor{lbmuted}{\bfseries No Experience (No Evolution)}} \\
Astra / xhigh & \lbpair{14.77}{8.50} & \lbpair{3.19}{0.00} & \lbpair{11.38}{5.00} & \lbpair{18.12}{0.00} & \lbpair{12.67}{10.00} & \lbpair{28.50}{27.50} \\
Astra / medium & \lbpair{6.58}{4.33} & \lbpair{1.81}{0.00} & \lbpair{2.88}{0.00} & \lbpair{5.88}{0.00} & \lbpair{7.33}{6.67} & \lbpair{15.00}{15.00} \\
Sol / xhigh & \lbpair{4.10}{3.17} & \lbpair{0.56}{0.00} & \lbpair{3.00}{2.50} & \lbpair{3.12}{0.00} & \lbpair{3.33}{3.33} & \lbpair{10.50}{10.00} \\
Luna / xhigh & \lbpair{0.08}{0.00} & \lbpair{0.42}{0.00} & \lbpair{0.00}{0.00} & \lbpair{0.00}{0.00} & \lbpair{0.00}{0.00} & \lbpair{0.00}{0.00} \\
\midrule
\multicolumn{7}{l}{\textcolor{lborange}{\bfseries Self-Evolved Harness}} \\
\rowcolor{lbours}
Astra / xhigh & \lbpair{\lbbest{38.64}}{\lbbest{30.17}} & \lbpair{\lbbest{22.36}}{\lbbest{16.67}} & \lbpair{\lbbest{29.88}}{\lbbest{15.00}} & \lbpair{\lbbest{29.50}}{\lbbest{12.50}} & \lbpair{\lbbest{59.33}}{\lbbest{56.67}} & \lbpair{\lbbest{52.12}}{\lbbest{50.00}} \\
Astra / medium & \lbpair{16.05}{8.67} & \lbpair{6.49}{0.00} & \lbpair{6.38}{0.00} & \lbpair{16.50}{0.00} & \lbpair{28.00}{23.33} & \lbpair{22.88}{20.00} \\
Sol / xhigh & \lbpair{4.09}{2.50} & \lbpair{0.28}{0.00} & \lbpair{0.38}{0.00} & \lbpair{12.12}{5.00} & \lbpair{0.17}{0.00} & \lbpair{7.50}{7.50} \\
Luna / xhigh & \lbpair{0.62}{0.00} & \lbpair{0.00}{0.00} & \lbpair{0.00}{0.00} & \lbpair{3.12}{0.00} & \lbpair{0.00}{0.00} & \lbpair{0.00}{0.00} \\
\midrule
\multicolumn{7}{l}{\textcolor{lbblue}{\bfseries Harness from Astra/xhigh}} \\
Astra / medium & \lbpair{30.60}{23.06} & \lbpair{17.67}{15.28} & \lbpair{11.38}{0.00} & \lbpair{26.12}{12.50} & \lbpair{55.83}{50.00} & \lbpair{42.00}{37.50} \\
Sol / xhigh & \lbpair{14.09}{12.03} & \lbpair{4.51}{3.47} & \lbpair{2.75}{0.00} & \lbpair{15.38}{10.00} & \lbpair{27.33}{26.67} & \lbpair{20.50}{20.00} \\
Luna / xhigh & \lbpair{6.55}{5.50} & \lbpair{0.21}{0.00} & \lbpair{0.38}{0.00} & \lbpair{4.50}{0.00} & \lbpair{20.17}{20.00} & \lbpair{7.50}{7.50} \\
\bottomrule
\end{tabularx}
\par\vspace{3pt}
\begin{minipage}{\linewidth}
\fontsize{7.5}{9}\selectfont
\textit{Notes:} Executor labels include the reasoning effort. Generalization variants, tasks within each category, and the five categories are equally weighted at their respective levels. Scene selection and evaluation details are summarized in Appendix~\ref{app:task_catalog}.
\end{minipage}
\endgroup
\end{table}

\textbf{Stronger experience acquisition benefits weaker executors.} Table~\ref{tab:paired-results} and Figure~\ref{fig:transfer} compare No Experience (No Evolution), Self-Evolved Harness, and Harness from Astra/xhigh at a fixed executor. Relative to No Experience, Astra/xhigh experience raises SR by 18.72 points for Astra/medium, 8.86 for Sol/xhigh, and 5.50 for Luna/xhigh. Sol reaches 12.03\% SR versus 2.50\% with its own harness; Luna reaches 5.50\% versus zero. Astra/medium tests a change in source reasoning effort, while Sol and Luna test a change in source model. These gains separate the ability to acquire useful experience from the ability to apply it: a weaker executor can use procedures and conditions that its own exploration failed to establish.

\begin{figure}[!htbp]
\centering
\includegraphics[width=\linewidth]{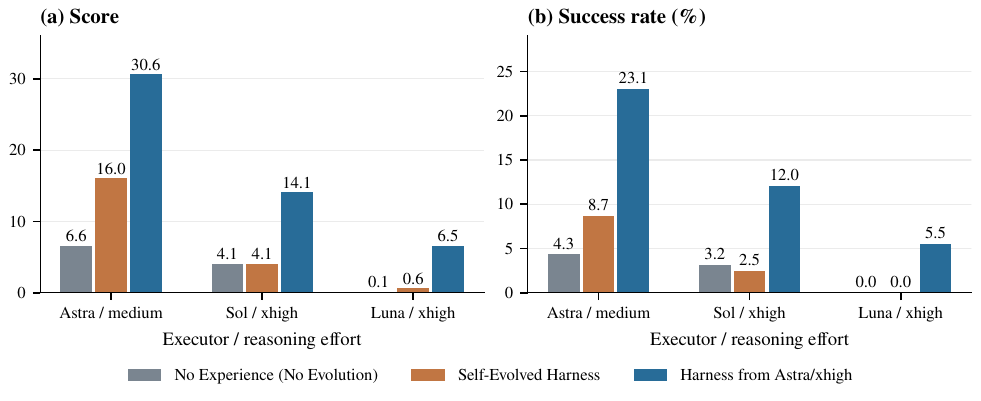}
\caption{\textbf{Experience transfer to weaker executors.} Each executor is evaluated on the same scenes under the three conditions used in Table~\ref{tab:paired-results}: No Experience (No Evolution), Self-Evolved Harness, and Harness from Astra/xhigh. The latter reuses frozen experience without additional exploration. Astra/medium compares source reasoning strengths within the same model; Sol/xhigh and Luna/xhigh test cross-model reuse.}
\label{fig:transfer}
\end{figure}

\textbf{Transferred experience supplies actionable conditions for control.} On \texttt{cover\_blocks}, Sol fails with its own harness after 680 native actions, but succeeds with Astra/xhigh's harness in 678 actions (Figure~\ref{fig:qualitative}b). Its own procedure loses a cup during lateral motion; the transferred harness distinguishes the grasp height for an empty cup from that for a cup covering a block. Sol applies these conditions and adapts the color order to the current scene. The case links successful reuse to selecting an appropriate operation and its place in the sequence, with nearly unchanged action expenditure.

\textbf{Other-task experience improves performance without target-task exploration.} On twelve fixed targets, related-task experience adds 14.18 Score points and 14.72 SR percentage points over No Experience; mismatched-task experience adds 18.65 and 16.67, respectively (Figure~\ref{fig:cross-task}). Same-task experience provides a reference gain of 24.43 Score points and 21.39 SR points. Both transfer settings reuse existing artifacts without exploring the target. Their gains are consistent with reuse of general control knowledge, including bounded motion, retention checks, and terminal return-home checks, that can guide execution across different tasks.

\begin{figure}[!htbp]
\centering
\includegraphics[width=\linewidth]{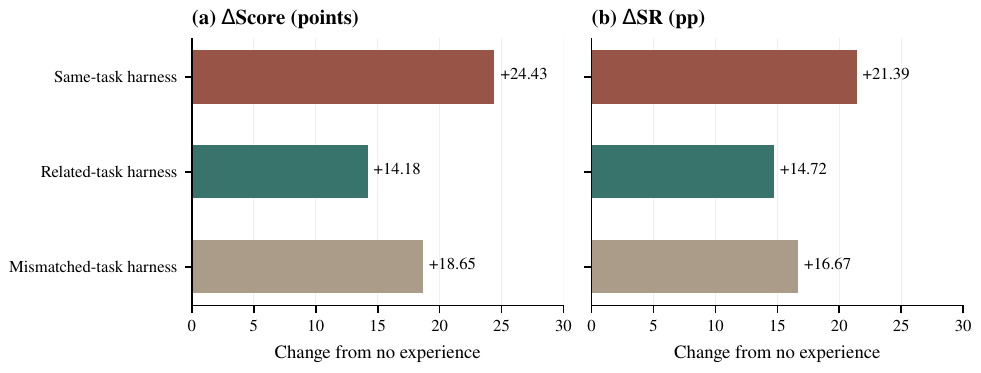}
\caption{\textbf{Performance gains from experience across tasks.} Each bar is the change from No Experience on the same target scenes, with Astra/xhigh fixed. $\Delta$Score uses points and $\Delta$SR uses percentage points; zero denotes the No Experience reference. Gains weight the twelve targets equally after combining Generalization variants. Related-task and mismatched-task harnesses require no new target-task exploration; the same-task harness is shown as a reference.}
\label{fig:cross-task}
\end{figure}

\textbf{Task-label similarity alone does not predict transfer benefit.} Mismatched transfer exceeds related transfer by 4.47 Score points and 1.94 SR points in this target set. The direction varies by task: on \texttt{swap\_T}, mismatched and related sources yield $\Delta$SR of +100 and 0 points; on \texttt{insert\_tubes}, they yield 0 and +40 points. The two mappings reuse the same source pool with equal aggregate file volume (Appendix~\ref{app:cross-task}). These reversals suggest that selecting useful experience requires inspecting the retained operations and their applicability conditions, beyond the semantic similarity of task labels.

\Needspace{6\baselineskip}
\subsection{How Experience Representation and Content Shape Reuse}
\label{sec:representation}

\begin{researchquestions}
\textbf{(1)} Can compact experience preserve the benefits of full interaction history? \textbf{(2)} Can an agent acquire useful knowledge without completing the task? \textbf{(3)} What makes retained knowledge useful for later execution?
\end{researchquestions}

To answer the questions above, we conduct a controlled comparison of Skills/Lessons and Raw History from the same Playground, using Astra/xhigh on matched Test scenes. We measure task performance and total token use, then inspect artifacts from unfinished tasks and link saved procedures and guidance to subsequent executions.

\textbf{Skills/Lessons achieve comparable task performance with fewer tokens.} Raw History provides the ordered programs and public feedback from the same Playground. It reaches 36.41 Score / 29.78\% SR, compared with 38.64 / 30.17\% for Skills/Lessons (Figure~\ref{fig:raw-history}). Mean total Test usage falls from 4.460M to 3.958M tokens per episode, a reduction of 11.3\%. Input usage drops from 4.445M to 3.939M, including an 87.9\% reduction in uncached input. Skills/Lessons organize exploration into reusable procedures and explicit guidance that the receiver can consult directly. The savings arise primarily on the input side, showing how this representation makes prior interaction more economical to use. Appendix~\ref{app:raw-history} gives the paired estimates and token accounting.

\begin{figure}[!htbp]
\centering
\includegraphics[width=0.7\linewidth]{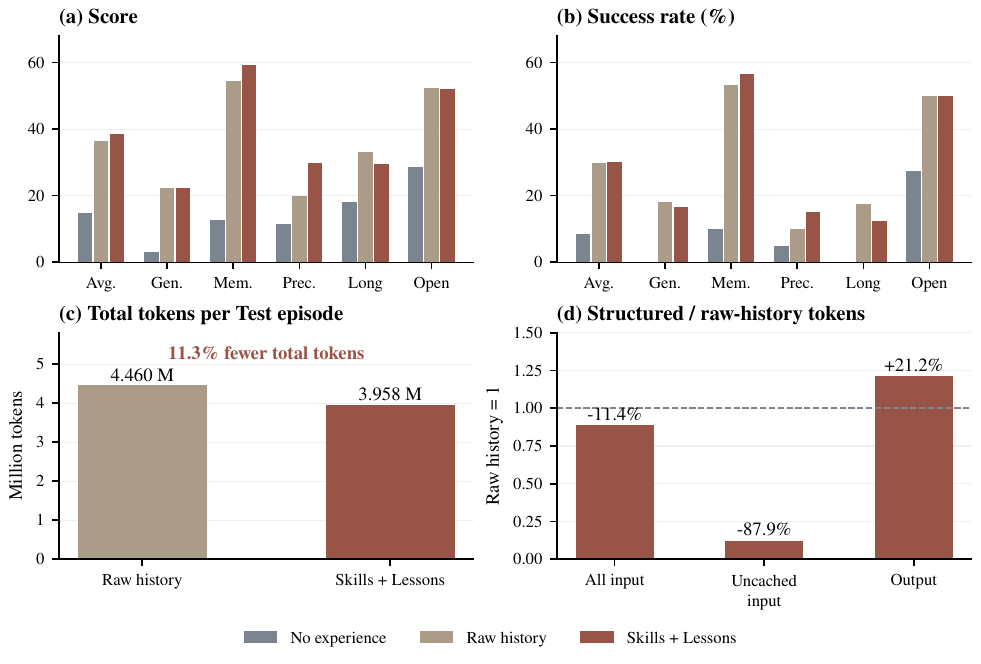}
\caption{\textbf{Comparable performance with lower token use.} The executor, source Playground, and Test scenes are fixed. (a--b) Raw History and Skills/Lessons achieve similar observed Score and SR. (c) Skills/Lessons reduce total Test tokens by 11.3\%. (d) The reduction comes from input usage, while output increases. Tokens are cumulative per episode, with cache reads included in input.}
\label{fig:raw-history}
\end{figure}

\textbf{Unfinished Playground tasks can still yield useful experience.} On \texttt{pick\_from\_conveyor\_by\_image}, Astra records no Playground success across 100 executions, yet its frozen harness raises Test Score from 0 to 40 and SR from 0 to 40\% (Table~\ref{tab:unfinished-experience}). The exploration establishes useful local operations: execution 74 completes a tracked grasp and test lift in 31 native actions. The retained files provide moving-target capture, contact-guarded transport, and guidance to update targets as the conveyor moves. These findings can be carried forward even when the entire task remains unfinished. Experience accumulates at the level of operations and their operating conditions, allowing later agents to build on partial progress.

% Generated by ccfa-workfiles/experiments/current-results/build_unfinished_experience.py.
\begin{table}[!htbp]
\centering\begingroup
% Shared typography and palette, matching the full-benchmark main table.
% Input inside a local group so the paper's body typography is unchanged.
\renewcommand{\rmdefault}{ptm}
\renewcommand{\sfdefault}{ptm}
\rmfamily
\captionsetup{font=small,skip=6pt}
\definecolor{lbblue}{HTML}{315F85}
\definecolor{lbteal}{HTML}{39746C}
\definecolor{lbviolet}{HTML}{6F598A}
\definecolor{lborange}{HTML}{9A683D}
\definecolor{lbmuted}{HTML}{59616A}
\definecolor{lbaverage}{HTML}{F1F5FA}
\definecolor{lbours}{HTML}{FAF4EE}
\newcommand{\lbbest}[1]{{\bfseries #1}}
\newcommand{\lbtype}[2]{\textcolor{#1}{\mbox{\fontsize{7.2}{9}\selectfont #2}}}
\newcommand{\lbhead}[1]{\makebox[50pt][c]{\bfseries #1}}
\newcommand{\lbpair}[2]{\makebox[23pt][r]{#1}\hspace{4pt}\makebox[23pt][r]{#2}}
\newcommand{\lbmetrics}{\lbpair{\textcolor{lbmuted}{\fontsize{7}{8}\selectfont Score}}{\textcolor{lbmuted}{\fontsize{7}{8}\selectfont SR (\%)}}}
\fontsize{8}{10}\selectfont
\setlength{\tabcolsep}{3.4pt}
\renewcommand{\arraystretch}{1.6}
\setlength{\heavyrulewidth}{0.65pt}
\setlength{\lightrulewidth}{0.35pt}
\setlength{\aboverulesep}{3pt}
\setlength{\belowrulesep}{3pt}

\small
\setlength{\tabcolsep}{6pt}
\renewcommand{\arraystretch}{1.15}
\begin{minipage}{0.82\linewidth}
\centering
\caption{\textbf{Useful experience without Playground task success.} Astra/xhigh on image-conditioned conveyor pickup; frozen experience is evaluated on new scenes.}
\label{tab:unfinished-experience}
\begin{tabularx}{\linewidth}{@{}p{0.23\linewidth}>{\raggedright\arraybackslash}X@{}}
\toprule
\textbf{Exploration} & \textbf{Observed progress and retained knowledge} \\
\midrule
Task outcome & No task success across 100 code executions. \\
Local progress & Execution 74: tracked grasp and 6\,cm lift in 31 native actions. \\
Skills & Track a moving target through closure; monitor contact during transport. \\
Lessons & Anticipate conveyor motion; verify object retention after a short lift. \\
\midrule
\textbf{Test condition} & \textbf{Score / SR (\%)} \\
No experience & 0 / 0 \\
\rowcolor{lbours}
Skills/Lessons & \textbf{40 / 40} \\
\bottomrule
\end{tabularx}
\end{minipage}
\endgroup
\end{table}

\textbf{Useful experience couples executable corrections with conditions and checks.} In the stacking trace, Test execution 9 reuses a controller with an 8\,mm translation bound; the agent checks the retained load and completes the stack at execution 15 (Figure~\ref{fig:experience-trace}). In covering, Astra validates a higher uncover grasp on three cups and records separate heights for empty-cup pickup and uncovering; Sol applies these conditions and succeeds at execution 17. These examples connect a concrete correction, a condition for applying it, and an observable check of its effect. Skills supply the operation, while Lessons guide its selection and verification in the current scene. Appendix~\ref{app:experience-content} provides the retained excerpts and reuse records.

\begin{figure}[!htbp]
\centering
\includegraphics[width=\linewidth]{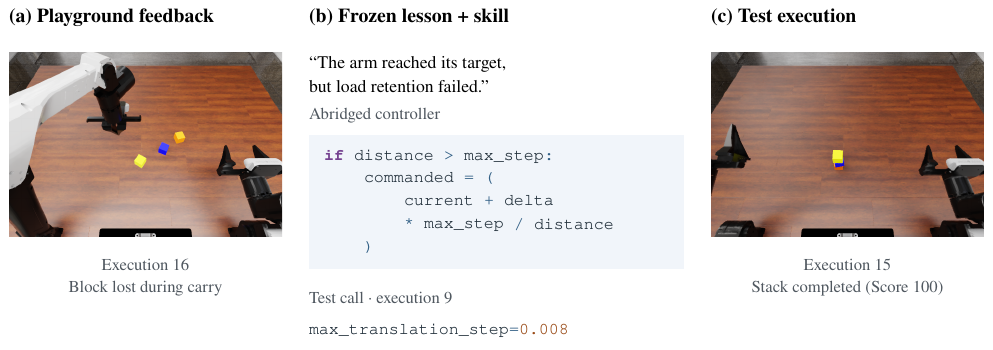}
\caption{\textbf{A correction retained with an execution check.} Left: Playground execution 16 records a dropped load. Center: the frozen lesson distinguishes reached pose from retained load, and the controller limits motion using the observed pose. Test execution 9 calls it with an 8\,mm bound. Right: the later successful stack at execution 15. The controller is abridged pseudocode; the lesson excerpt and call parameter come from the retained artifacts and Test program.}
\label{fig:experience-trace}
\end{figure}

\Needspace{6\baselineskip}
\subsection{Performance Gains and the Cost of Reusing Experience}
\label{sec:cost-reuse}

\begin{researchquestions}
\textbf{(1)} What execution cost accompanies the gains from experience? \textbf{(2)} Can structuring experience lower the cost of reuse?
\end{researchquestions}

To answer the questions above, we compare No Experience, Skills/Lessons, and Raw History with Astra/xhigh on matched scenes. Skills/Lessons and Raw History share the same Playground source. We report per-episode model expense and interaction counts, including failures, using a fixed token-rate schedule and excluding Playground acquisition.

\textbf{Experience improves performance with higher execution expense.} Score rises from 14.77 to 38.64 and SR from 8.50\% to 30.17\%, while mean total Test cost increases from \$4.058 to \$5.732 per episode (+41.3\%; Table~\ref{tab:cost-results}). The executor consults retained procedures and guidance while continuing to observe, submit code, and check feedback. Experience therefore improves completion while still requiring sustained interaction with the environment.

% Generated by cost_analysis.py from audited matched Astra/xhigh results.
\begin{table}[!htbp]
\centering\begingroup
% Shared typography and palette, matching the full-benchmark main table.
% Input inside a local group so the paper's body typography is unchanged.
\renewcommand{\rmdefault}{ptm}
\renewcommand{\sfdefault}{ptm}
\rmfamily
\captionsetup{font=small,skip=6pt}
\definecolor{lbblue}{HTML}{315F85}
\definecolor{lbteal}{HTML}{39746C}
\definecolor{lbviolet}{HTML}{6F598A}
\definecolor{lborange}{HTML}{9A683D}
\definecolor{lbmuted}{HTML}{59616A}
\definecolor{lbaverage}{HTML}{F1F5FA}
\definecolor{lbours}{HTML}{FAF4EE}
\newcommand{\lbbest}[1]{{\bfseries #1}}
\newcommand{\lbtype}[2]{\textcolor{#1}{\mbox{\fontsize{7.2}{9}\selectfont #2}}}
\newcommand{\lbhead}[1]{\makebox[50pt][c]{\bfseries #1}}
\newcommand{\lbpair}[2]{\makebox[23pt][r]{#1}\hspace{4pt}\makebox[23pt][r]{#2}}
\newcommand{\lbmetrics}{\lbpair{\textcolor{lbmuted}{\fontsize{7}{8}\selectfont Score}}{\textcolor{lbmuted}{\fontsize{7}{8}\selectfont SR (\%)}}}
\fontsize{8}{10}\selectfont
\setlength{\tabcolsep}{3.4pt}
\renewcommand{\arraystretch}{1.6}
\setlength{\heavyrulewidth}{0.65pt}
\setlength{\lightrulewidth}{0.35pt}
\setlength{\aboverulesep}{3pt}
\setlength{\belowrulesep}{3pt}

\small\setlength{\tabcolsep}{6pt}\renewcommand{\arraystretch}{1.15}
\caption{\textbf{Performance and execution expense at a fixed executor.} Astra/xhigh on matched dev scenes. Costs and interaction counts are per-episode means, including failures. Raw History and Skills/Lessons use the same Playground source.}
\label{tab:cost-results}
\begin{tabularx}{\linewidth}{@{}>{\raggedright\arraybackslash}X rr >{\columncolor{lbaverage}}r rr@{}}
\toprule
\textbf{Experience} & \textbf{Score} & \textbf{SR (\%)} & \shortstack[r]{\bfseries Cost\\\bfseries (USD)} & \shortstack[r]{\bfseries Code\\\bfseries executions} & \shortstack[r]{\bfseries Native\\\bfseries actions} \\
\midrule
No Experience & 14.77 & 8.50 & 4.058 & 22.3 & 723.0 \\
Raw History & 36.41 & 29.78 & 12.203 & 23.7 & 705.1 \\
\rowcolor{lbours}
Skills/Lessons & 38.64 & 30.17 & 5.732 & 29.0 & 722.6 \\
\bottomrule
\end{tabularx}
% Leave extra room before the following section heading.
\par\vspace{13pt}
\endgroup\end{table}

\Needspace{3\baselineskip}
\textbf{Structured experience is cheaper to reuse than raw history.} Using the same Playground source, Skills/Lessons cost \$5.732 per Test episode versus \$12.203 for Raw History, a 53.0\% reduction, with SRs of 30.17\% and 29.78\%, respectively. Skills/Lessons retain reusable procedures and conditions for applying them, reducing the input processed when consulting prior exploration. This compact representation preserves similar observed performance at a lower total execution cost.

\section{Discussion}
\label{sec:discussion}

\paragraph{Does embodied control necessarily require a VLA?}
EmbodiedEvo controls robots through code and achieves higher average scores and success rates than the evaluated VLA, WAM, and Agent+VLA systems on RoboDojo-Sim. These results prompt us to reconsider the role of agents in embodied intelligence: beyond high-level planning, agents can directly generate low level robot motion commands and refine control programs through physical feedback.

\paragraph{Precision manipulation remains a major bottleneck.}
Our Full-set Precision success is only 13.25\%; tube insertion achieves higher intermediate scores but still fails. Possible causes include inaccurate contact geometry, unstable transitions between actions, and insufficient validation of local procedures. Future work could use a few demonstrations to guide poses and action sequences, then focus exploration on difficult contact stages. Repeated execution under perturbed initial conditions could help validate reliability before retaining procedures for reuse. Reinforcement learning could further improve exploration and recovery, turning interaction into more reliable, reusable contact experience.

\paragraph{Harness evolution expands what fixed primitives can accomplish.}
With the underlying primitives held fixed, the agent combines observation, motion, and gripper control into reusable procedures. Skills organize these combinations into parameterized programs with execution checks, while Lessons provide applicability conditions and recovery guidance. The harness thus forms an evolving control abstraction that lets subsequent agents adapt existing operations to the current scene. Embodied capability therefore depends on both the available primitives and how agents compose, validate, and reuse them.

% \paragraph{Autonomous interaction offers a path to further improvement.}
% Beyond our current frozen-model approach, scalable reinforcement learning could improve the coding agent's exploration, program generation, and recovery decisions, reducing dependence on human demonstrations. Evaluation should extend beyond RoboDojo-Sim to additional benchmarks and robot platforms. Long agent-call latency also motivates reusing validated routines with fewer model calls and assessing both success and elapsed task time.

\section{Conclusion}
\label{sec:conclusion}

We presented EmbodiedEvo, a framework for coding agents to turn physical interaction into reusable Skills and Lessons without updating model parameters. On RoboDojo-Sim, EmbodiedEvo achieves 44.48 Average Score and 37.82\% SR. Paired evaluations show consistent gains across different input settings. Skills and Lessons achieve comparable observed performance to raw interaction history at 53.0\% lower Test execution cost. These findings suggest that harness evolution improves the starting point of task solving by preserving executable operations and the conditions for applying them. Precise contact and reliable long-horizon execution remain important directions for extending this approach.

\par\FloatBarrier
\endgroup

% ── References ──────────────────────────────────────────────────────────────

\bibliographystyle{unsrtnat}
\begingroup
\renewcommand{\bibfont}{\fontsize{8.5}{10}\selectfont}
\setlength{\bibsep}{1pt plus 0.5pt}
\bibliography{ref}
\endgroup

% ── Appendix ─────────────────────────────────────────────────────────────────

\clearpage
\appendix

\begingroup
\compactfloatlayout
% Verbatim excerpts from the public Workspace templates at revision
% 0687b1414a4dcbc6677fe4efde69fd649c90a48f (verified 2026-10-14).
% Source hashes and retained ranges are in instruction-excerpts/manifest.json.
\section{Playground and Test Instructions}
\label{app:instructions}

\begingroup
\tcbset{instruction excerpt/.style={
    enhanced, breakable,
    colback=brandbg, colframe=brandprimary,
    colbacktitle=brandprimary, coltitle=brandfg,
    fonttitle=\bfseries,
    arc=1mm, boxrule=0.5pt,
    left=8pt, right=8pt, top=5pt, bottom=5pt,
    toptitle=3pt, bottomtitle=3pt,
    before skip=8pt, after skip=8pt
}}
\lstdefinestyle{instructiontext}{
    language={},
    basicstyle=\fontfamily{pcr}\fontsize{8.5}{10.2}\selectfont,
    morecomment=[l]{\#}, commentstyle=\bfseries\color{brandaccent},
    columns=fullflexible, keepspaces=true,
    breaklines=true, breakatwhitespace=true, breakindent=0pt,
    showstringspaces=false, numbers=none,
    aboveskip=0pt, belowskip=0pt,
    frame=none, upquote=true, emptylines=0
}

\begin{tcolorbox}[instruction excerpt,title={Playground instruction}]
\begin{lstlisting}[style=instructiontext]
# Goal

You are an embodied coding agent controlling a robot through code. You are currently in Playground Phase. Your goal is to explore the current robot task, develop a working solution, and extract reusable skills and lessons that may transfer to future scenes.

[...]

Organize your work as an iterative loop:
observe -> reason -> implement -> record reusable knowledge when it appears -> submit -> inspect -> revise

[...]

Whenever an observation reveals a reusable controller pattern, a repeated failure, a recovery strategy, or an important edge case, pause and write it to `skills/` or `lessons/` immediately while the evidence is fresh. Do not wait until the end of the session to record it.

# Environment and Primitives

Read `primitives/primitives.md` before interacting with the environment. You have four native primitives:

- `get_instruction()`: read the current task and goal.
- `get_observation()`: read the latest observation, including camera images and robot state.
- `step(action, ...)`: execute a native controller action.
- `reset()`: restore the current scene and renew its official native action allowance. The total {{EXECUTION_BUDGET}} code-execution requests and elapsed time are never reset.

[...]

# Evolution: skills and lessons

[...]

A skill should capture a general behavior or procedure, not merely rename a sequence of open-loop `step` calls. A useful skill should:

- be parameterized by relevant targets, tolerances, and control budgets;
- derive scene-dependent decisions from observations and explicit inputs;
- be reusable across similar tasks or scenes;
- specify its preconditions and limitations;
- include evidence from experiments that supports its behavior.

Lessons are textual knowledge about what worked, what failed, and how to recover. Record useful:

- failure patterns;
- diagnostic signals;
- recovery strategies;
- important edge cases;
- unresolved questions.

[...]

# Playground rules

You may reset with `reset()` in code or `python scripts/env.py reset` and update `skills/` and `lessons/`. Exploration uses the configured single scene. Success here is evidence about this scene; transfer to unseen scenes remains unverified.

Keep the task, primitives, and success conditions unchanged. Your transferable deliverables are the skills and lessons you develop.

You cannot use `reset()` in formal evaluation.

[...]

When done, out of slots, or unable to continue, review the current `skills/` and `lessons/` and run `finish`. If the connection breaks, query `status` before resubmitting.
\end{lstlisting}
\end{tcolorbox}

\clearpage
\begin{tcolorbox}[instruction excerpt,title={Test instruction}]
\begin{lstlisting}[style=instructiontext]
# Goal

You are an embodied coding agent controlling a robot through code. You are currently in Test Phase. Your goal is to complete the current robot task using the available primitives and the skills and lessons from Playground.

[...]

Organize your work as an iterative loop:
observe -> reason -> implement -> submit -> inspect -> revise

[...]

# Environment and Primitives

Read `primitives/primitives.md` before interacting with the environment. You have three native primitives:

- `get_instruction()`: read the current task and goal.
- `get_observation()`: read the latest observation, including camera images and robot state.
- `step(action, ...)`: execute a native controller action.

[...]

# Using skills and lessons

[...]

When using a skill:

- check its preconditions, limitations, and experimental evidence against the current observations;
- choose targets, tolerances, and control budgets appropriate to the current task;
- derive scene-dependent decisions from live observations and explicit inputs;
- inspect feedback after execution and decide the next stage from the resulting robot state.

Use lessons to recognize failure patterns, interpret diagnostic signals, and choose recovery strategies. Distinguish verified findings, scene-specific observations, and untested hypotheses; a result from Playground may not transfer to this scene.

[...]

# Test rules

Work within a single continuous episode. You may observe, revise code, execute stages, and recover from failures using the current physical state. Both `reset()` in code and `python scripts/env.py reset` are forbidden in Test.

Keep the task, primitives, success conditions, skills, and lessons unchanged. Complete the task using the frozen knowledge and the feedback from this scene.

Each Test scene starts with a fresh agent, workspace, and simulator session. Only frozen skills and lessons carry over from Playground; its executed program, temporary files, and observations do not. This scene's code, observations, and analysis do not carry over to other Test scenes.

[...]

Reset is forbidden in Test. All **{{EXECUTION_BUDGET}} code-execution requests** share the same **{{NATIVE_ACTION_LIMIT}} native action steps** for this scene; neither allowance can be restored.

[...]

When the task succeeds, you run out of execution slots, or you cannot continue, save episode-specific work in `submission/` and `scratch/` and run `finish`. The environment determines success. If the connection breaks, query `status` before resubmitting.
\end{lstlisting}
\end{tcolorbox}
\endgroup

\clearpage

\section{Per-Task Results}
\label{app:task_catalog}

Tables~\ref{tab:task-results} and~\ref{tab:task-results-full} report complete dev- and full-set Score and SR. The dev table compares Astra and EmbodiedEvo with and without wiki on matched scenes; the full table reports EmbodiedEvo with wiki. Generalization averages standard and random variants equally, and Average weights the five task categories equally.

% Generated by write_per_task_results.py from frozen audited results.
\begin{table}[H]
\centering\begingroup
% Shared typography and palette, matching the full-benchmark main table.
% Input inside a local group so the paper's body typography is unchanged.
\renewcommand{\rmdefault}{ptm}
\renewcommand{\sfdefault}{ptm}
\rmfamily
\captionsetup{font=small,skip=6pt}
\definecolor{lbblue}{HTML}{315F85}
\definecolor{lbteal}{HTML}{39746C}
\definecolor{lbviolet}{HTML}{6F598A}
\definecolor{lborange}{HTML}{9A683D}
\definecolor{lbmuted}{HTML}{59616A}
\definecolor{lbaverage}{HTML}{F1F5FA}
\definecolor{lbours}{HTML}{FAF4EE}
\newcommand{\lbbest}[1]{{\bfseries #1}}
\newcommand{\lbtype}[2]{\textcolor{#1}{\mbox{\fontsize{7.2}{9}\selectfont #2}}}
\newcommand{\lbhead}[1]{\makebox[50pt][c]{\bfseries #1}}
\newcommand{\lbpair}[2]{\makebox[23pt][r]{#1}\hspace{4pt}\makebox[23pt][r]{#2}}
\newcommand{\lbmetrics}{\lbpair{\textcolor{lbmuted}{\fontsize{7}{8}\selectfont Score}}{\textcolor{lbmuted}{\fontsize{7}{8}\selectfont SR (\%)}}}
\fontsize{8}{10}\selectfont
\setlength{\tabcolsep}{3.4pt}
\renewcommand{\arraystretch}{1.6}
\setlength{\heavyrulewidth}{0.65pt}
\setlength{\lightrulewidth}{0.35pt}
\setlength{\aboverulesep}{3pt}
\setlength{\belowrulesep}{3pt}

\fontsize{8.5}{10}\selectfont
\setlength{\tabcolsep}{4pt}\renewcommand{\arraystretch}{1.06}
\caption{\textbf{Per-task dev-set results.} Each cell is Score / SR (\%); all four conditions use Astra/xhigh on matched scenes. Shaded category rows report means; the final row reports the overall Average.}\label{tab:task-results}
\begin{tabularx}{\linewidth}{>{\raggedright\arraybackslash}Xrrrr}
\toprule
\multirow{2}{*}{\bfseries Task} & \multicolumn{2}{c}{\bfseries Task description} & \multicolumn{2}{c}{\bfseries Task description + wiki} \\
\cmidrule(lr){2-3}\cmidrule(lr){4-5}
& Astra & EmbodiedEvo & Astra & EmbodiedEvo \\
\midrule
\rowcolor{lbaverage}
\textcolor{lbblue}{\bfseries Generalization} & \lbbest{3.19 / 0.00} & \lbbest{22.36 / 16.67} & \lbbest{12.33 / 7.64} & \lbbest{36.15 / 31.25} \\
arrange\_largest\_number & 3.75 / 0.00 & 13.75 / 0.00 & 8.33 / 0.00 & 49.17 / 41.67 \\
fold\_clothes & 11.67 / 0.00 & 13.33 / 0.00 & 33.33 / 16.67 & 70.00 / 66.67 \\
hang\_mugs & 0.00 / 0.00 & 10.00 / 0.00 & 0.00 / 0.00 & 8.75 / 0.00 \\
make\_toast & 0.00 / 0.00 & 0.00 / 0.00 & 0.00 / 0.00 & 0.00 / 0.00 \\
pack\_objects\_into\_box & 6.67 / 0.00 & 22.50 / 0.00 & 6.67 / 0.00 & 22.50 / 0.00 \\
pour\_liquid\_into\_cup & 0.00 / 0.00 & 0.00 / 0.00 & 0.00 / 0.00 & 41.67 / 41.67 \\
push\_T & 0.00 / 0.00 & 83.33 / 83.33 & 0.00 / 0.00 & 83.33 / 83.33 \\
sort\_nesting\_dolls\_by\_size & 0.00 / 0.00 & 0.00 / 0.00 & 0.00 / 0.00 & 25.00 / 25.00 \\
stack\_blocks & 10.00 / 0.00 & 100.00 / 100.00 & 50.42 / 41.67 & 100.00 / 100.00 \\
stack\_bowls & 6.25 / 0.00 & 25.42 / 16.67 & 40.83 / 33.33 & 26.67 / 16.67 \\
store\_laptop\_and\_headphones & 0.00 / 0.00 & 0.00 / 0.00 & 8.33 / 0.00 & 6.67 / 0.00 \\
sweep\_blocks & 0.00 / 0.00 & 0.00 / 0.00 & 0.00 / 0.00 & 0.00 / 0.00 \\
\rowcolor{lbaverage}
\textcolor{lbblue}{\bfseries Memory} & \lbbest{12.67 / 10.00} & \lbbest{59.33 / 56.67} & \lbbest{39.67 / 36.67} & \lbbest{70.33 / 66.67} \\
cover\_blocks & 6.00 / 0.00 & 86.00 / 80.00 & 60.00 / 60.00 & 100.00 / 100.00 \\
imitate\_sorting\_sequence & 10.00 / 0.00 & 90.00 / 80.00 & 38.00 / 20.00 & 62.00 / 40.00 \\
match\_and\_pick\_from\_conveyor & 60.00 / 60.00 & 100.00 / 100.00 & 100.00 / 100.00 & 100.00 / 100.00 \\
press\_by\_number & 0.00 / 0.00 & 0.00 / 0.00 & 20.00 / 20.00 & 100.00 / 100.00 \\
swap\_T & 0.00 / 0.00 & 80.00 / 80.00 & 20.00 / 20.00 & 60.00 / 60.00 \\
swap\_blocks & 0.00 / 0.00 & 0.00 / 0.00 & 0.00 / 0.00 & 0.00 / 0.00 \\
\rowcolor{lbaverage}
\textcolor{lbblue}{\bfseries Precision} & \lbbest{11.38 / 5.00} & \lbbest{29.88 / 15.00} & \lbbest{15.00 / 7.50} & \lbbest{23.12 / 12.50} \\
build\_tower & 2.00 / 0.00 & 26.00 / 0.00 & 6.00 / 0.00 & 10.00 / 0.00 \\
deposit\_coin & 8.00 / 0.00 & 12.00 / 0.00 & 8.00 / 0.00 & 36.00 / 20.00 \\
fasten\_screws & 28.00 / 0.00 & 70.00 / 40.00 & 34.00 / 0.00 & 64.00 / 40.00 \\
insert\_key & 9.00 / 0.00 & 15.00 / 0.00 & 12.00 / 0.00 & 15.00 / 0.00 \\
insert\_tubes & 4.00 / 0.00 & 36.00 / 0.00 & 0.00 / 0.00 & 40.00 / 20.00 \\
play\_Xylophone & 20.00 / 20.00 & 40.00 / 40.00 & 40.00 / 40.00 & 20.00 / 20.00 \\
plug\_in\_charger & 0.00 / 0.00 & 40.00 / 40.00 & 0.00 / 0.00 & 0.00 / 0.00 \\
pour\_balls\_into\_vase & 20.00 / 20.00 & 0.00 / 0.00 & 20.00 / 20.00 & 0.00 / 0.00 \\
\rowcolor{lbaverage}
\textcolor{lbblue}{\bfseries Long-Horizon} & \lbbest{18.12 / 0.00} & \lbbest{29.50 / 12.50} & \lbbest{19.38 / 0.00} & \lbbest{33.75 / 20.00} \\
classify\_objects & 22.00 / 0.00 & 40.00 / 0.00 & 30.00 / 0.00 & 47.00 / 20.00 \\
fill\_egg\_holder & 7.00 / 0.00 & 7.00 / 0.00 & 7.00 / 0.00 & 20.00 / 0.00 \\
fill\_pen\_holder & 15.00 / 0.00 & 5.00 / 0.00 & 8.00 / 0.00 & 2.00 / 0.00 \\
make\_kong & 0.00 / 0.00 & 0.00 / 0.00 & 0.00 / 0.00 & 0.00 / 0.00 \\
organize\_table & 15.00 / 0.00 & 40.00 / 0.00 & 40.00 / 0.00 & 30.00 / 0.00 \\
play\_stacking\_toy & 6.00 / 0.00 & 4.00 / 0.00 & 18.00 / 0.00 & 10.00 / 0.00 \\
play\_tic\_tac\_toe & 55.00 / 0.00 & 100.00 / 100.00 & 32.00 / 0.00 & 100.00 / 100.00 \\
put\_bottles\_into\_dustbin & 25.00 / 0.00 & 40.00 / 0.00 & 20.00 / 0.00 & 61.00 / 40.00 \\
\rowcolor{lbaverage}
\textcolor{lbblue}{\bfseries Open} & \lbbest{28.50 / 27.50} & \lbbest{52.12 / 50.00} & \lbbest{17.00 / 15.00} & \lbbest{48.75 / 45.00} \\
align\_blocks & 60.00 / 60.00 & 80.00 / 80.00 & 20.00 / 20.00 & 100.00 / 100.00 \\
classify\_objects\_by\_language & 24.00 / 20.00 & 48.00 / 40.00 & 16.00 / 0.00 & 46.00 / 20.00 \\
general\_pickup & 80.00 / 80.00 & 60.00 / 60.00 & 40.00 / 40.00 & 80.00 / 80.00 \\
pick\_from\_conveyor\_by\_image & 0.00 / 0.00 & 40.00 / 40.00 & 20.00 / 20.00 & 20.00 / 20.00 \\
pour\_by\_language & 0.00 / 0.00 & 0.00 / 0.00 & 0.00 / 0.00 & 0.00 / 0.00 \\
solve\_equation & 20.00 / 20.00 & 100.00 / 100.00 & 20.00 / 20.00 & 80.00 / 80.00 \\
stack\_blocks\_by\_language & 44.00 / 40.00 & 84.00 / 80.00 & 20.00 / 20.00 & 64.00 / 60.00 \\
store\_tools\_in\_toolbox & 0.00 / 0.00 & 5.00 / 0.00 & 0.00 / 0.00 & 0.00 / 0.00 \\
\midrule
\rowcolor{lbours}
\textbf{Average} & \lbbest{14.77 / 8.50} & \lbbest{38.64 / 30.17} & \lbbest{20.67 / 13.36} & \lbbest{42.42 / 35.08} \\
\bottomrule
\end{tabularx}
\endgroup\end{table}

% Generated by write_per_task_results.py from frozen audited results.
\begin{table}[p]
\centering\begingroup
% Shared typography and palette, matching the full-benchmark main table.
% Input inside a local group so the paper's body typography is unchanged.
\renewcommand{\rmdefault}{ptm}
\renewcommand{\sfdefault}{ptm}
\rmfamily
\captionsetup{font=small,skip=6pt}
\definecolor{lbblue}{HTML}{315F85}
\definecolor{lbteal}{HTML}{39746C}
\definecolor{lbviolet}{HTML}{6F598A}
\definecolor{lborange}{HTML}{9A683D}
\definecolor{lbmuted}{HTML}{59616A}
\definecolor{lbaverage}{HTML}{F1F5FA}
\definecolor{lbours}{HTML}{FAF4EE}
\newcommand{\lbbest}[1]{{\bfseries #1}}
\newcommand{\lbtype}[2]{\textcolor{#1}{\mbox{\fontsize{7.2}{9}\selectfont #2}}}
\newcommand{\lbhead}[1]{\makebox[50pt][c]{\bfseries #1}}
\newcommand{\lbpair}[2]{\makebox[23pt][r]{#1}\hspace{4pt}\makebox[23pt][r]{#2}}
\newcommand{\lbmetrics}{\lbpair{\textcolor{lbmuted}{\fontsize{7}{8}\selectfont Score}}{\textcolor{lbmuted}{\fontsize{7}{8}\selectfont SR (\%)}}}
\fontsize{8}{10}\selectfont
\setlength{\tabcolsep}{3.4pt}
\renewcommand{\arraystretch}{1.6}
\setlength{\heavyrulewidth}{0.65pt}
\setlength{\lightrulewidth}{0.35pt}
\setlength{\aboverulesep}{3pt}
\setlength{\belowrulesep}{3pt}

\fontsize{8.5}{10}\selectfont
\setlength{\tabcolsep}{6pt}\renewcommand{\arraystretch}{1.13}
\begin{minipage}{0.90\linewidth}
\centering
\caption{\textbf{Per-task full-set results for EmbodiedEvo (Astra/xhigh).} Task description + wiki, from the completed evaluation in Table~\ref{tab:full-results}. Shaded category rows report means; the final row reports the overall Average.}
\label{tab:task-results-full}
\begin{tabularx}{\linewidth}{>{\raggedright\arraybackslash}X >{\raggedleft\arraybackslash}p{50pt} >{\raggedleft\arraybackslash}p{50pt}}
\toprule
\textbf{Task / category} & \textbf{Score} & \textbf{SR (\%)} \\
\midrule
\rowcolor{lbaverage}
\textcolor{lbblue}{\bfseries Generalization} & \lbbest{34.12} & \lbbest{29.67} \\
arrange\_largest\_number & 58.20 & 46.00 \\
fold\_clothes & 66.80 & 60.00 \\
hang\_mugs & 5.40 & 0.00 \\
make\_toast & 1.50 & 0.00 \\
pack\_objects\_into\_box & 20.30 & 4.00 \\
pour\_liquid\_into\_cup & 28.00 & 28.00 \\
push\_T & 86.00 & 86.00 \\
sort\_nesting\_dolls\_by\_size & 8.00 & 8.00 \\
stack\_blocks & 71.60 & 68.00 \\
stack\_bowls & 54.50 & 50.00 \\
store\_laptop\_and\_headphones & 3.20 & 0.00 \\
sweep\_blocks & 6.00 & 6.00 \\
\addlinespace[3pt]
\rowcolor{lbaverage}
\textcolor{lbblue}{\bfseries Memory} & \lbbest{71.90} & \lbbest{69.67} \\
cover\_blocks & 100.00 & 100.00 \\
imitate\_sorting\_sequence & 71.40 & 58.00 \\
match\_and\_pick\_from\_conveyor & 76.00 & 76.00 \\
press\_by\_number & 98.00 & 98.00 \\
swap\_T & 86.00 & 86.00 \\
swap\_blocks & 0.00 & 0.00 \\
\addlinespace[3pt]
\rowcolor{lbaverage}
\textcolor{lbblue}{\bfseries Precision} & \lbbest{23.65} & \lbbest{13.25} \\
build\_tower & 8.60 & 0.00 \\
deposit\_coin & 37.20 & 22.00 \\
fasten\_screws & 60.00 & 38.00 \\
insert\_key & 15.00 & 0.00 \\
insert\_tubes & 36.40 & 14.00 \\
play\_Xylophone & 16.00 & 16.00 \\
plug\_in\_charger & 4.00 & 4.00 \\
pour\_balls\_into\_vase & 12.00 & 12.00 \\
\addlinespace[3pt]
\rowcolor{lbaverage}
\textcolor{lbblue}{\bfseries Long-Horizon} & \lbbest{37.20} & \lbbest{23.50} \\
classify\_objects & 44.30 & 22.00 \\
fill\_egg\_holder & 13.80 & 0.00 \\
fill\_pen\_holder & 2.00 & 0.00 \\
make\_kong & 0.00 & 0.00 \\
organize\_table & 41.50 & 2.00 \\
play\_stacking\_toy & 18.80 & 0.00 \\
play\_tic\_tac\_toe & 100.00 & 100.00 \\
put\_bottles\_into\_dustbin & 77.20 & 64.00 \\
\addlinespace[3pt]
\rowcolor{lbaverage}
\textcolor{lbblue}{\bfseries Open} & \lbbest{55.52} & \lbbest{53.00} \\
align\_blocks & 98.00 & 98.00 \\
classify\_objects\_by\_language & 63.40 & 46.00 \\
general\_pickup & 86.00 & 86.00 \\
pick\_from\_conveyor\_by\_image & 38.00 & 38.00 \\
pour\_by\_language & 0.40 & 0.00 \\
solve\_equation & 76.00 & 76.00 \\
stack\_blocks\_by\_language & 82.40 & 80.00 \\
store\_tools\_in\_toolbox & 0.00 & 0.00 \\
\midrule
\rowcolor{lbours}
\textbf{Average} & \lbbest{44.48} & \lbbest{37.82} \\
\bottomrule
\end{tabularx}
\par\vspace{4pt}
\end{minipage}\endgroup
\end{table}

\FloatBarrier
\section{Cross-Task Transfer}
\label{app:cross-task}

Table~\ref{tab:cross-task} gives the source assignments and outcomes for the twelve targets in Section~\ref{sec:experience-transfer}. Related assignments pair tasks with similar operations; the mismatched assignment permutes the same source pool. Both maps were fixed before execution and expose 146 files totaling 388,507 bytes, counted once per target package. Only read-only Skills/Lessons are transferred; the instruction, goal, and scene remain those of the target. Neither transfer condition performs target-task exploration.

All conditions use Astra/xhigh on the same target scenes. Generalization variants receive equal weight within each target, then the twelve targets receive equal weight. The table retains absolute outcomes so that the gains over No Experience in Figure~\ref{fig:cross-task} can be reconstructed; these subset means are separate from the full-benchmark Average.

% Fixed target subset; not the full benchmark Average.
\begin{table}[!htbp]
\centering\begingroup
% Shared typography and palette, matching the full-benchmark main table.
% Input inside a local group so the paper's body typography is unchanged.
\renewcommand{\rmdefault}{ptm}
\renewcommand{\sfdefault}{ptm}
\rmfamily
\captionsetup{font=small,skip=6pt}
\definecolor{lbblue}{HTML}{315F85}
\definecolor{lbteal}{HTML}{39746C}
\definecolor{lbviolet}{HTML}{6F598A}
\definecolor{lborange}{HTML}{9A683D}
\definecolor{lbmuted}{HTML}{59616A}
\definecolor{lbaverage}{HTML}{F1F5FA}
\definecolor{lbours}{HTML}{FAF4EE}
\newcommand{\lbbest}[1]{{\bfseries #1}}
\newcommand{\lbtype}[2]{\textcolor{#1}{\mbox{\fontsize{7.2}{9}\selectfont #2}}}
\newcommand{\lbhead}[1]{\makebox[50pt][c]{\bfseries #1}}
\newcommand{\lbpair}[2]{\makebox[23pt][r]{#1}\hspace{4pt}\makebox[23pt][r]{#2}}
\newcommand{\lbmetrics}{\lbpair{\textcolor{lbmuted}{\fontsize{7}{8}\selectfont Score}}{\textcolor{lbmuted}{\fontsize{7}{8}\selectfont SR (\%)}}}
\fontsize{8}{10}\selectfont
\setlength{\tabcolsep}{3.4pt}
\renewcommand{\arraystretch}{1.6}
\setlength{\heavyrulewidth}{0.65pt}
\setlength{\lightrulewidth}{0.35pt}
\setlength{\aboverulesep}{3pt}
\setlength{\belowrulesep}{3pt}

\caption{\textbf{Cross-task sources and target outcomes.} Each outcome is Score / SR (\%). Related and mismatched sources are fixed before these transfer runs; B and D reuse the same target-scene results.}
\label{tab:cross-task}
\setlength{\tabcolsep}{2pt}
\begin{tabularx}{\linewidth}{*{3}{>{\raggedright\arraybackslash}X} *{4}{c}}
\toprule
\bfseries Target & \bfseries Related source & \bfseries Mismatched source & \bfseries B & \bfseries D & \bfseries Related & \bfseries Mismatched \\
& & & \lbmetrics & \lbmetrics & \lbmetrics & \lbmetrics \\
\midrule
stack bowls & stack blocks & swap blocks & \lbpair{6.25}{0.00} & \lbpair{25.42}{16.67} & \lbpair{25.42}{16.67} & \lbpair{15.00}{0.00} \\
stack blocks & stack bowls & swap T & \lbpair{10.00}{0.00} & \lbpair{100.00}{100.00} & \lbpair{100.00}{100.00} & \lbpair{83.33}{83.33} \\
insert tubes & plug in charger & classify objects & \lbpair{4.00}{0.00} & \lbpair{36.00}{0.00} & \lbpair{60.00}{40.00} & \lbpair{4.00}{0.00} \\
plug in charger & insert tubes & classify objects by language & \lbpair{0.00}{0.00} & \lbpair{40.00}{40.00} & \lbpair{0.00}{0.00} & \lbpair{0.00}{0.00} \\
fill pen holder & fill egg holder & pour liquid into cup & \lbpair{15.00}{0.00} & \lbpair{5.00}{0.00} & \lbpair{2.00}{0.00} & \lbpair{17.00}{0.00} \\
fill egg holder & fill pen holder & pour by language & \lbpair{7.00}{0.00} & \lbpair{7.00}{0.00} & \lbpair{8.00}{0.00} & \lbpair{14.00}{0.00} \\
swap blocks & swap T & stack bowls & \lbpair{0.00}{0.00} & \lbpair{0.00}{0.00} & \lbpair{0.00}{0.00} & \lbpair{0.00}{0.00} \\
swap T & swap blocks & stack blocks & \lbpair{0.00}{0.00} & \lbpair{80.00}{80.00} & \lbpair{0.00}{0.00} & \lbpair{100.00}{100.00} \\
classify objects & classify objects by language & insert tubes & \lbpair{22.00}{0.00} & \lbpair{40.00}{0.00} & \lbpair{39.00}{20.00} & \lbpair{30.00}{0.00} \\
classify objects by language & classify objects & plug in charger & \lbpair{24.00}{20.00} & \lbpair{48.00}{40.00} & \lbpair{24.00}{20.00} & \lbpair{32.00}{20.00} \\
pour liquid into cup & pour by language & fill pen holder & \lbpair{0.00}{0.00} & \lbpair{0.00}{0.00} & \lbpair{0.00}{0.00} & \lbpair{16.67}{16.67} \\
pour by language & pour liquid into cup & fill egg holder & \lbpair{0.00}{0.00} & \lbpair{0.00}{0.00} & \lbpair{0.00}{0.00} & \lbpair{0.00}{0.00} \\
\midrule
\multicolumn{3}{l}{\bfseries Fixed-target mean} & \lbpair{7.35}{1.67} & \lbpair{31.78}{23.06} & \lbpair{21.53}{16.39} & \lbpair{26.00}{18.33} \\
\bottomrule
\end{tabularx}
\par\vspace{3pt}\begin{minipage}{\linewidth}\fontsize{7.5}{9}\selectfont
\textit{Notes:} B: No Experience; D: same-task harness. Standard/random variants are averaged equally within each Generalization task, then the twelve target tasks receive equal weight. These subset means are separate from the full benchmark Average. All executors and experience producers are Astra/xhigh. No new Playground is run for either transfer condition.
\end{minipage}\endgroup
\end{table}

\FloatBarrier
\section{Supporting Case Evidence}

\subsection{Capability Changes}
\label{app:capability-cases}

Table~\ref{tab:capability-evidence} summarizes the execution records behind Section~\ref{sec:capability}. All cases use Astra/xhigh; paired comparisons use the same scene. The evidence bundle retains the original programs, feedback, file reads, and camera frames. These selected cases illustrate mechanisms, not failure frequencies or isolated causal effects.

\begin{table}[!htbp]
\centering
\small
\setlength{\tabcolsep}{5pt}
\renewcommand{\arraystretch}{1.14}
\caption{\textbf{Execution evidence for the capability analysis.} Action counts refer to native simulator actions; layout IDs identify the matched scenes.}
\label{tab:capability-evidence}
\begin{tabularx}{\linewidth}{@{}p{0.21\linewidth}X@{}}
\toprule
\textbf{Task / layout} & \textbf{Key execution evidence} \\
\midrule
Tic-tac-toe / 27 & Astra first retains a ring at action 367 and ends with Score 50 at the 1,100-action limit. EmbodiedEvo includes \texttt{rim\_pick.py}, retains a ring at action 38, and succeeds at action 919. First rings and arms differ. \\
\addlinespace
Imitation sorting / 4 & Astra moves before the demonstration finishes and terminates at action 256 with Score 0; feedback does not identify the cause. EmbodiedEvo observes the new order through action 455 and succeeds at action 1,585. \\
\addlinespace
Insert key / 44 & EmbodiedEvo's receiver closes empty in execution 14 and remains unsecured through execution 21. Donor release reaches the 300-action limit with Score 15. Failure occurs during handover, before insertion. \\
\bottomrule
\end{tabularx}
\end{table}

\subsection{Experience Reuse and Its Limits}
\label{app:experience-content}

Figure~\ref{fig:qualitative} compares successful reuse, cross-model transfer, and a regression. Figure~\ref{fig:skills-lessons} provides the complete retained grasp function and two Lessons used in the successful stacking trace in Appendix~\ref{app:agent-trace}.

\begin{figure}[!htbp]
\centering
\includegraphics[width=\linewidth]{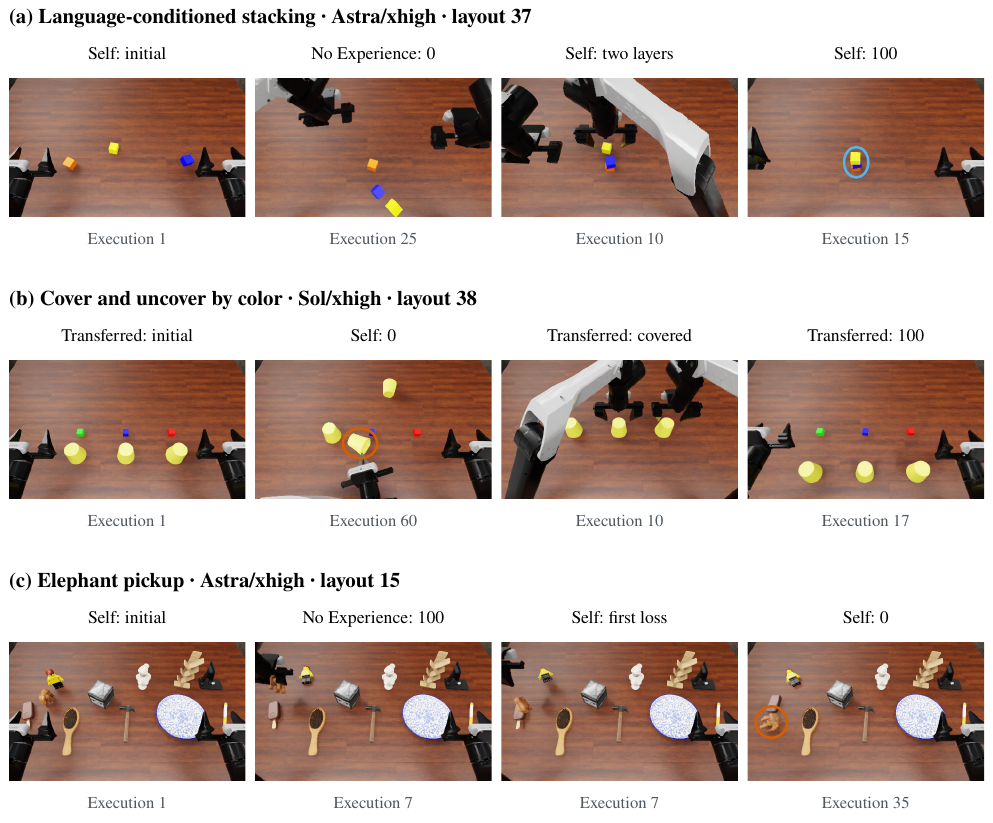}
\caption{\textbf{Success, transfer, and regression on matched scenes.} (a) Language stacking, Astra/xhigh, layout 37: the self-harness completes the stack. (b) Cover/uncover, Sol/xhigh, layout 38: Astra's harness succeeds where Sol's own harness fails. (c) Elephant pickup, Astra/xhigh, layout 15: Astra succeeds without experience but loses the object with its self-harness. Frames are original observations from independent runs, with identical crops. Labels show official Scores and execution indices.}
\label{fig:qualitative}
\end{figure}

\paragraph{Stacking: a correction with a check.}
The frozen lesson records, ``The arm reached its target, but load retention failed.'' Test execution 2 includes \texttt{ee\_move.py} and \texttt{pick\_from\_above.py} verbatim. Execution 9 uses an 8\,mm translation bound after visual retention checking; execution 15 completes the stack. This establishes actual reuse, without isolating the bound's causal contribution.

\paragraph{Covering: conditions that transfer.}
Sol's own artifact describes a ``pickup-only experimental skill, not a validated carry or place controller''; its Test loses the cup during transport. Astra's lesson instead distinguishes empty-cup pickup from a higher uncover grasp, validated on three cups. Sol uses these separate heights in executions 2, 11, and 13 and succeeds in execution 17.

\paragraph{Pickup: a transfer boundary.}
Astra's saved retracting lift succeeds on scissors in Playground. On the elephant, Test execution 7 reports \texttt{pose\_reached}, but the following observation shows the object was lost. Source-scene success therefore leaves retention on new geometry unresolved; the trace does not isolate the saved advice as the cause of failure.

\begin{figure}[!htbp]
\centering
\begingroup
\renewcommand{\rmdefault}{ptm}\renewcommand{\sfdefault}{ptm}\rmfamily
\noindent\makebox[\linewidth][l]{\bfseries (a) Skill: a guarded, budgeted grasp function}\par\vspace{3pt}
\begin{lstlisting}[style=experience, basicstyle=\fontfamily{pcr}\fontsize{7.5}{9}\selectfont]
def pick_from_above(arm, grasp_pose, lift_z, action_budget, grip_dwell=12, descend_step=0.010, lift_step=0.008, alignment_tolerance=0.003):
    """Pick from an already aligned hover; return for visual retention verification."""
    obs = get_observation()
    current = obs['state'][arm + '_ee_pose']
    lateral_error = sum((current[j] - grasp_pose[j]) ** 2 for j in range(2)) ** 0.5
    if lateral_error > alignment_tolerance or current[2] < grasp_pose[2]:
        return obs, {'error': 'requires_aligned_hover', 'steps': 0}
    used = 0
    stages = [('descend', list(grasp_pose), 1.0, 3, descend_step),
              ('close', list(grasp_pose), 0.0, grip_dwell, descend_step)]
    raised = list(grasp_pose)
    raised[2] = lift_z
    stages.append(('lift', raised, 0.0, 3, lift_step))
    for name, target, grip, dwell, speed in stages:
        remaining = action_budget - used
        if remaining < dwell:
            return obs, {'error': 'budget_exhausted', 'stage': name, 'steps': used}
        obs, result = move_ee(arm, target, grip, max_steps=remaining, min_steps=dwell, max_translation_step=speed)
        used += result['steps']
        if not result.get('reached', False):
            return obs, {'stage': name, 'steps': used, 'motion': result}
    return obs, {'stage': 'lifted_needs_visual_check', 'steps': used, 'gripper': obs['state'][arm + '_ee_joint_state']}
\end{lstlisting}
\par\vspace{7pt}
\begin{minipage}[t]{0.48\linewidth}
\textbf{(b) Lesson: verify retention, then carry slowly}\par\vspace{3pt}
\fontsize{8.5}{10.5}\selectfont
A Playground carry dropped the block even though the arm reached its target. The saved rule is to bound each translation and inspect the load after each carry stage. Test execution 9 uses an 8\,mm bound after the agent inspects the lifted blue block.
\begin{lstlisting}[style=experience, firstline=3,lastline=4]
obs, result = move_ee('right', base_xy+[1.045]+down, 0.0, max_steps=46, max_translation_step=0.008)
print(result)
\end{lstlisting}
\end{minipage}\hfill
\begin{minipage}[t]{0.48\linewidth}
\textbf{(c) Lesson: reserve actions for the terminal check}\par\vspace{3pt}
\fontsize{8.5}{10.5}\selectfont
The saved rule requires home commands after placement and immediate stopping on termination. The stable Test stack still has success=false at execution 14. Execution 15 triggers success after 10 home actions, with five actions left.
\begin{lstlisting}[style=experience, firstline=3,lastline=7]
for i in range(15):
    obs, reward, terminated, truncated, info = step(home)
    if terminated or truncated:
        print({'steps':i+1,'reward':reward,'terminated':terminated,'truncated':truncated,'success':info.get('success',False)})
        break
\end{lstlisting}
\end{minipage}
\endgroup
\caption{\textbf{A retained Skill and two Lessons in actual reuse.} (a) The complete frozen \texttt{pick\_from\_above} function composes the measured-pose \texttt{move\_ee} controller into descent, closure, and lifting. It checks alignment and the action budget, then returns for visual retention verification. The same source is included verbatim in Test execution 2 and called for orange, blue, and yellow in executions 4, 8, and 12. (b--c) The lesson descriptions paraphrase the frozen \texttt{downward\_approach.md}; the highlighted Test snippets are verbatim. The complete episode is shown in Appendix~\ref{app:agent-trace}.}
\label{fig:skills-lessons}
\end{figure}

\section{Raw-History Comparison and Cost Accounting}
\label{app:raw-history}
\label{app:raw_data}

\paragraph{Matched comparison.}
Raw History supplies the same Playground's ordered programs and public feedback, including raw observations and failed attempts, through read-only files. It excludes Skills/Lessons and private scores and is not summarized by another model. Both representations use Astra/xhigh on matched Test scenes, with no reset. Skills/Lessons exceed Raw History by 2.23 Score points (95\% interval: $-2.60$--7.25) and 0.39 SR points ($-5.11$--5.94). Intervals use 10,000 paired task-bootstrap draws within categories (seed 20261006). Similar observed performance is not a formal equivalence result.

\paragraph{Tokens and expense.}
Figure~\ref{fig:raw-history} and Table~\ref{tab:cost-results} average terminal API-usage counters over complete Test episodes, including failures and repeated context; Playground acquisition is excluded. Total tokens equal input plus output, with cache reads included in input and reasoning included in output. A missing final event in one local Raw History trace is recovered from its terminal report, without imputation. Cache reads account for 82.6\% of Raw History input and 97.6\% of Skills/Lessons input. The fixed Astra rates are \$10/\$1/\$12.5/\$50 per million ordinary-input/cache-read/cache-write/output tokens. These nominal rates explain the larger cost saving (53.0\%) than total-token saving (11.3\%); the amounts are estimates, not invoices.

\FloatBarrier
% Generated by build_agent_trace.py from the saved public episode evidence.
\Needspace{6\baselineskip}
\section{Complete Successful Agent Trace}
\label{app:agent-trace}

This trace covers every accepted execution in the Astra/xhigh self-harness run of \texttt{stack\_blocks\_by\_language}, public layout 37. The episode makes 15 execution requests and uses 395 of 400 native actions, ending with official Score 100 and success=true. Each movement stage below pairs a camera image that the agent actually opened \emph{before} the displayed action with its submitted decision comment, highlighted code, and subsequent feedback. Images retain their original field of view. Execution 1 retrieves the instruction alongside the initial image read; executions 2--15 advance the same simulation session without reset.

\paragraph{What the agent reads.}
The agent reads the fixed Test delegation, the native primitive manual, the Skills index, both \texttt{ee\_move} and \texttt{pick\_from\_above} source files and usage notes, and \texttt{lessons/downward\_approach.md}. It also reads the initial state metadata and opens the head-camera image. The session records these accesses at lines 18, 25, and 32. The frozen source is prepended to execution 2; later calls reuse the persistent Python namespace. The second helper is reproduced in Figure~\ref{fig:skills-lessons}. Execution 7 adds an episode-local \texttt{move\_both} helper to reposition the empty arms together.

\begin{quote}\small
\textbf{Task returned by} \texttt{get\_instruction()}: ``stack the blocks from bottom to top in the order of orange, blue, and yellow, then reset the robot arm.''
\end{quote}
Here ``reset the robot arm'' means commanding the saved starting joints, as execution 15 does with \texttt{step(home)}. The Test episode never calls the environment's \texttt{reset()} primitive.

\begin{figure}[H]
\centering
\begin{minipage}[t]{0.32\linewidth}\centering
\includegraphics[width=\linewidth]{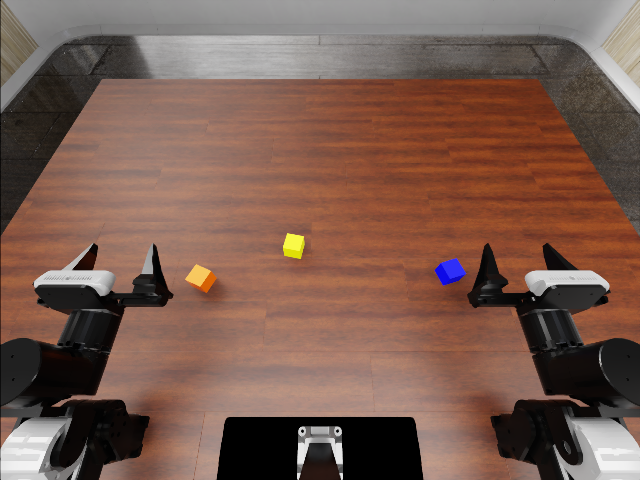}\par\vspace{6pt}
\small Initial observation: opened before control
\end{minipage}\hspace{0.06\linewidth}
\begin{minipage}[t]{0.32\linewidth}\centering
\includegraphics[width=\linewidth]{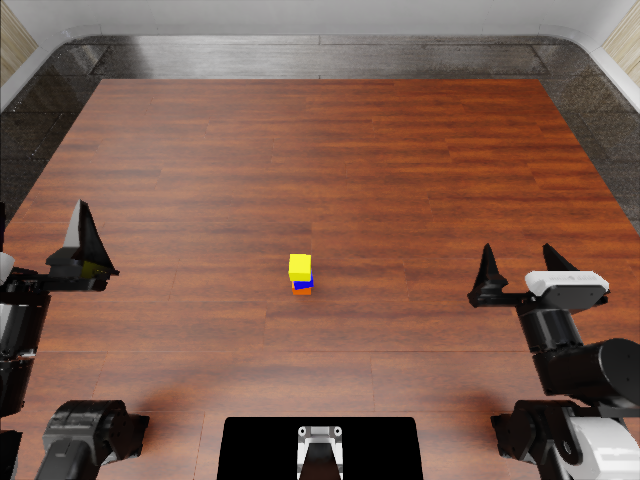}\par\vspace{6pt}
\small Terminal observation: recorded after execution 15
\end{minipage}
\caption{\textbf{Endpoints of the complete stacking episode.} The initial head view is opened by the agent; the terminal image is recorded by the environment. After execution 15, the agent receives the success feedback and ends without a further image-opening call. The required final order is orange--blue--yellow from bottom to top.}
\label{fig:trace-endpoints}
\end{figure}

\paragraph{Reading the sequence.}
Decision text is copied from the agent's submitted code comments; execution 1 has an editorial description of its instruction call. The comments are the recorded rationale for an action, not an independent annotation of what an image proves. Feedback text summarizes the subsequent stdout and public execution counters. The code panels show the complete stage-specific action code except for the repeated imported helpers in execution 2 and the local helper definition in execution 7. The source bundle retains all submitted programs, full stdout, frozen helpers and lessons, and the image-access index in \texttt{agent-trace/trace.json}.
\Needspace{12\baselineskip}
\subsection*{Execution sequence}
\begingroup\setlength{\parskip}{2pt}
\noindent\begin{minipage}[t]{\linewidth}
{\small\bfseries Execution 1: Read the task}\hfill{\footnotesize +0 actions; 0/400 used}\par\vspace{4pt}\nointerlineskip
\begin{minipage}[t]{0.27\linewidth}\vspace{0pt}\centering
\includegraphics[width=\linewidth]{ccfa-workfiles/experiments/current-results/agent-trace/images/00-head.png}\par\vspace{3pt}
{\fontsize{8}{10}\selectfont Head camera\par Observation 0, during instruction retrieval}\end{minipage}\hfill
\begin{minipage}[t]{0.70\linewidth}\vspace{0pt}\fontsize{8.5}{10.5}\selectfont\raggedright
\textbf{Recorded decision.} Retrieve the current task instruction before issuing a robot action.\par
\begin{lstlisting}[style=experience,firstline=1,lastline=1]
print(get_instruction())
\end{lstlisting}
\textbf{Feedback.} The instruction specifies orange, blue, then yellow; no native action is used.
\end{minipage}\par\vspace{3pt}\noindent{\color{black!15}\rule{\linewidth}{0.35pt}}\par\vspace{3pt}\end{minipage}\par
\noindent\begin{minipage}[t]{\linewidth}
{\small\bfseries Execution 2: Load the skills and approach orange}\hfill{\footnotesize +22 actions; 22/400 used}\par\vspace{4pt}\nointerlineskip
\begin{minipage}[t]{0.27\linewidth}\vspace{0pt}\centering
\includegraphics[width=\linewidth]{ccfa-workfiles/experiments/current-results/agent-trace/images/00-head.png}\par\vspace{3pt}
{\fontsize{8}{10}\selectfont Head camera\par Observation 0, before execution~2}\end{minipage}\hfill
\begin{minipage}[t]{0.70\linewidth}\vspace{0pt}\fontsize{8.5}{10.5}\selectfont\raggedright
\textbf{Recorded decision.} Orange must form the base, with blue and then yellow above it, before returning home. Save home joints and approach orange at a raised pose to calibrate its location safely.\par
\begin{lstlisting}[style=experience,firstline=62,lastline=67]
obs = get_observation()
home = {k:list(v) for k,v in obs['state'].items() if k.endswith('joint_state')}
down = [0.5,-0.5,0.5,0.5]
obs, result = move_ee('left', [-0.20,-0.10,1.10]+down, 1.0, max_steps=45)
print(result)
print(obs['state'])
\end{lstlisting}
\textbf{Feedback.} The raised target is reached in 22 actions; the initial home joint dictionary is saved.
\end{minipage}\par\vspace{3pt}\noindent{\color{black!15}\rule{\linewidth}{0.35pt}}\par\vspace{3pt}\end{minipage}\par
\noindent\begin{minipage}[t]{\linewidth}
{\small\bfseries Execution 3: Correct orange alignment}\hfill{\footnotesize +7 actions; 29/400 used}\par\vspace{4pt}\nointerlineskip
\begin{minipage}[t]{0.27\linewidth}\vspace{0pt}\centering
\includegraphics[width=\linewidth]{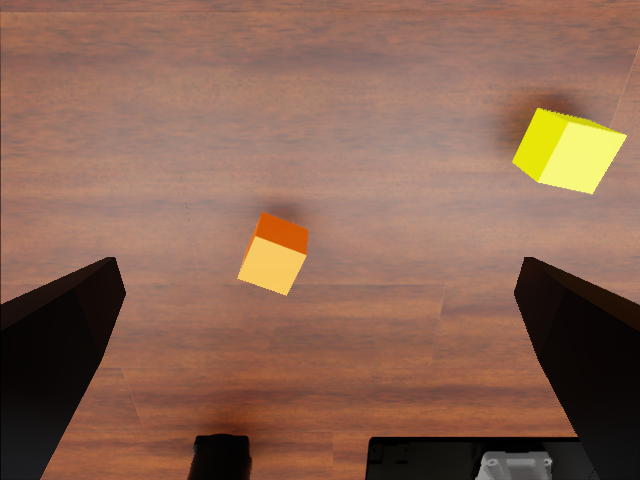}\par\vspace{3pt}
{\fontsize{8}{10}\selectfont Left wrist camera\par Observation 2, before execution~3}\end{minipage}\hfill
\begin{minipage}[t]{0.70\linewidth}\vspace{0pt}\fontsize{8.5}{10.5}\selectfont\raggedright
\textbf{Recorded decision.} Orange is left and forward of the wrist's calibrated grasp alignment at this height. Shift the raised left gripper toward it before descending.\par
\begin{lstlisting}[style=experience,firstline=3,lastline=4]
obs, result = move_ee('left', [-0.225,-0.17,1.10]+down, 1.0, max_steps=12)
print(result)
\end{lstlisting}
\textbf{Feedback.} The raised alignment target is reached in 7 actions.
\end{minipage}\par\vspace{3pt}\noindent{\color{black!15}\rule{\linewidth}{0.35pt}}\par\vspace{3pt}\end{minipage}\par
\noindent\begin{minipage}[t]{\linewidth}
{\small\bfseries Execution 4: Grasp and lift orange}\hfill{\footnotesize +55 actions; 84/400 used}\par\vspace{4pt}\nointerlineskip
\begin{minipage}[t]{0.27\linewidth}\vspace{0pt}\centering
\includegraphics[width=\linewidth]{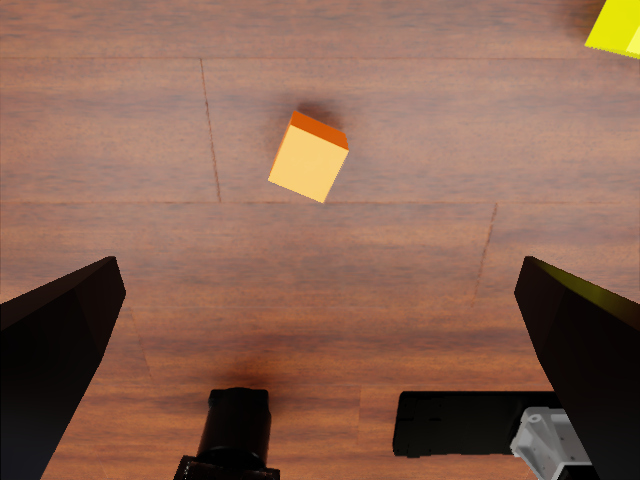}\par\vspace{3pt}
{\fontsize{8}{10}\selectfont Left wrist camera\par Observation 3, before execution~4}\end{minipage}\hfill
\begin{minipage}[t]{0.70\linewidth}\vspace{0pt}\fontsize{8.5}{10.5}\selectfont\raggedright
\textbf{Recorded decision.} The raised wrist view localizes orange near (-0.231, -0.188). Align precisely, then use the validated low grasp and slow vertical lift before checking retention.\par
\begin{lstlisting}[style=experience,firstline=3,lastline=8]
obs, result = move_ee('left', [-0.231,-0.188,1.10]+down, 1.0, max_steps=6)
print('alignment', result)
if result.get('reached', False):
    obs, result = pick_from_above('left', [-0.231,-0.188,0.925]+down, 1.065, action_budget=52)
    print('pick', result)
print(obs['state']['left_ee_pose'])
\end{lstlisting}
\textbf{Feedback.} Alignment uses 3 actions and the grasp uses 52. The skill returns lifted\_needs\_visual\_check.
\end{minipage}\par\vspace{3pt}\noindent{\color{black!15}\rule{\linewidth}{0.35pt}}\par\vspace{3pt}\end{minipage}\par
\noindent\begin{minipage}[t]{\linewidth}
{\small\bfseries Execution 5: Carry the retained orange block}\hfill{\footnotesize +28 actions; 112/400 used}\par\vspace{4pt}\nointerlineskip
\begin{minipage}[t]{0.27\linewidth}\vspace{0pt}\centering
\includegraphics[width=\linewidth]{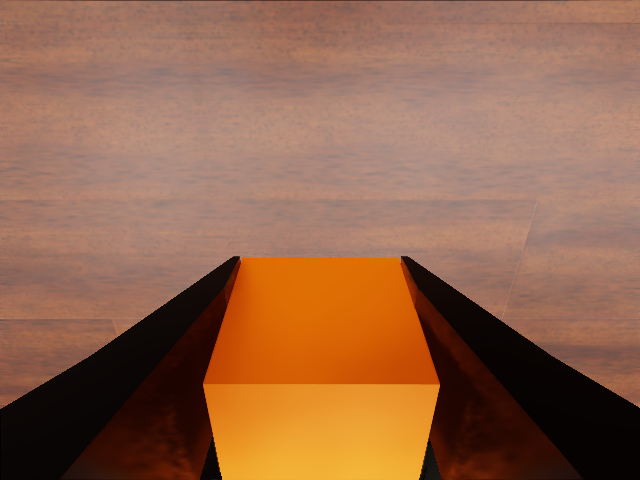}\par\vspace{3pt}
{\fontsize{8}{10}\selectfont Left wrist camera\par Observation 4, before execution~5}\end{minipage}\hfill
\begin{minipage}[t]{0.70\linewidth}\vspace{0pt}\fontsize{8.5}{10.5}\selectfont\raggedright
\textbf{Recorded decision.} Orange is visibly retained between the left fingers after lifting. Carry it slowly toward a central base location while keeping clearance above the table.\par
\begin{lstlisting}[style=experience,firstline=3,lastline=6]
base_xy = [-0.035,-0.188]
obs, result = move_ee('left', base_xy+[1.045]+down, 0.0, max_steps=34, max_translation_step=0.008)
print(result)
print(obs['state']['left_ee_pose'])
\end{lstlisting}
\textbf{Feedback.} The 8 mm translation bound reaches the central base position in 28 actions.
\end{minipage}\par\vspace{3pt}\noindent{\color{black!15}\rule{\linewidth}{0.35pt}}\par\vspace{3pt}\end{minipage}\par
\noindent\begin{minipage}[t]{\linewidth}
{\small\bfseries Execution 6: Place the base and retreat}\hfill{\footnotesize +35 actions; 147/400 used}\par\vspace{4pt}\nointerlineskip
\begin{minipage}[t]{0.27\linewidth}\vspace{0pt}\centering
\includegraphics[width=\linewidth]{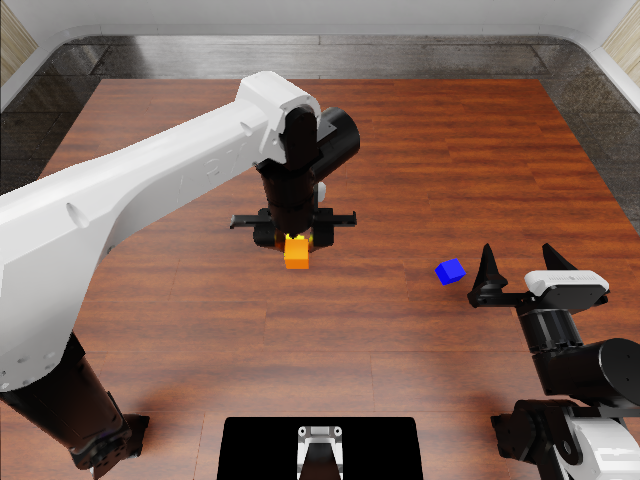}\par\vspace{3pt}
{\fontsize{8}{10}\selectfont Head camera\par Observation 5, before execution~6}\end{minipage}\hfill
\begin{minipage}[t]{0.70\linewidth}\vspace{0pt}\fontsize{8.5}{10.5}\selectfont\raggedright
\textbf{Recorded decision.} Orange remains held at the central location with clear table space below. Lower it to table height, release with a dwell, then withdraw vertically.\par
\begin{lstlisting}[style=experience,firstline=3,lastline=10]
obs, result = move_ee('left', base_xy+[0.928]+down, 0.0, max_steps=23, max_translation_step=0.008)
print('lower', result)
if result.get('reached', False):
    obs, result = move_ee('left', base_xy+[0.928]+down, 1.0, max_steps=8, min_steps=8)
    print('release', result)
    if not result.get('terminated', False) and not result.get('truncated', False):
        obs, result = move_ee('left', base_xy+[1.065]+down, 1.0, max_steps=15, max_translation_step=0.018)
        print('retreat', result)
\end{lstlisting}
\textbf{Feedback.} Descent, release, and retreat take 17, 8, and 10 actions, respectively.
\end{minipage}\par\vspace{3pt}\noindent{\color{black!15}\rule{\linewidth}{0.35pt}}\par\vspace{3pt}\end{minipage}\par
\noindent\begin{minipage}[t]{\linewidth}
{\small\bfseries Execution 7: Clear the left arm and approach blue}\hfill{\footnotesize +12 actions; 159/400 used}\par\vspace{4pt}\nointerlineskip
\begin{minipage}[t]{0.27\linewidth}\vspace{0pt}\centering
\includegraphics[width=\linewidth]{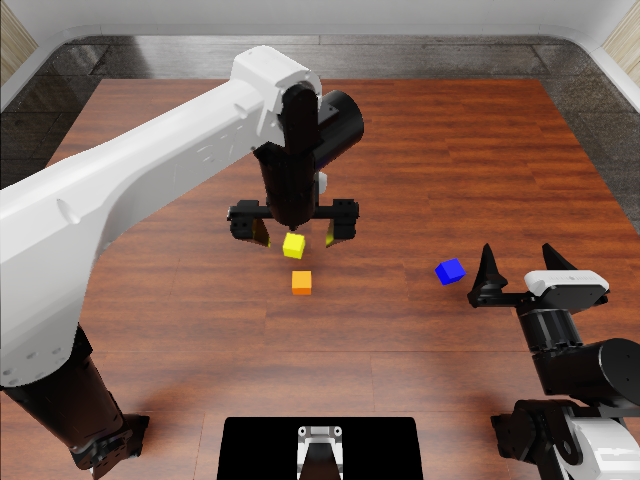}\par\vspace{3pt}
{\fontsize{8}{10}\selectfont Head camera\par Observation 6, before execution~7}\end{minipage}\hfill
\begin{minipage}[t]{0.70\linewidth}\vspace{0pt}\fontsize{8.5}{10.5}\selectfont\raggedright
\textbf{Recorded decision.} Orange is stable on the table, and blue is next in the stack. Move the empty left arm aside while the right arm approaches blue at safe clearance.\par
\begin{lstlisting}[style=experience,firstline=22,lastline=24]
obs, result = move_both([-0.24,-0.24,1.065]+down, [0.25,-0.16,1.065]+down, max_steps=22)
print(result)
print(obs['state']['right_ee_pose'])
\end{lstlisting}
\textbf{Feedback.} A locally defined move\_both helper reaches both targets in 12 actions.
\end{minipage}\par\vspace{3pt}\noindent{\color{black!15}\rule{\linewidth}{0.35pt}}\par\vspace{3pt}\end{minipage}\par
\noindent\begin{minipage}[t]{\linewidth}
{\small\bfseries Execution 8: Grasp and lift blue}\hfill{\footnotesize +48 actions; 207/400 used}\par\vspace{4pt}\nointerlineskip
\begin{minipage}[t]{0.27\linewidth}\vspace{0pt}\centering
\includegraphics[width=\linewidth]{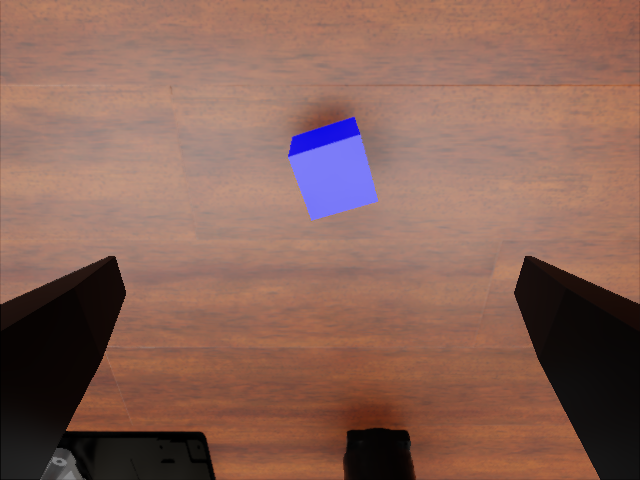}\par\vspace{3pt}
{\fontsize{8}{10}\selectfont Right wrist camera\par Observation 7, before execution~8}\end{minipage}\hfill
\begin{minipage}[t]{0.70\linewidth}\vspace{0pt}\fontsize{8.5}{10.5}\selectfont\raggedright
\textbf{Recorded decision.} Blue is slightly right and forward of the calibrated wrist alignment. Correct the raised pose, grasp at table height, and lift slowly to verify retention.\par
\begin{lstlisting}[style=experience,firstline=3,lastline=8]
blue_xy = [0.2575,-0.166]
obs, result = move_ee('right', blue_xy+[1.065]+down, 1.0, max_steps=5)
print('align',result)
if result.get('reached',False):
    obs, result = pick_from_above('right', blue_xy+[0.925]+down, 1.045, action_budget=48)
    print('pick',result)
\end{lstlisting}
\textbf{Feedback.} Alignment takes 3 actions and the inherited grasp takes 45; visual verification remains required.
\end{minipage}\par\vspace{3pt}\noindent{\color{black!15}\rule{\linewidth}{0.35pt}}\par\vspace{3pt}\end{minipage}\par
\noindent\begin{minipage}[t]{\linewidth}
{\small\bfseries Execution 9: Carry blue above orange}\hfill{\footnotesize +41 actions; 248/400 used}\par\vspace{4pt}\nointerlineskip
\begin{minipage}[t]{0.27\linewidth}\vspace{0pt}\centering
\includegraphics[width=\linewidth]{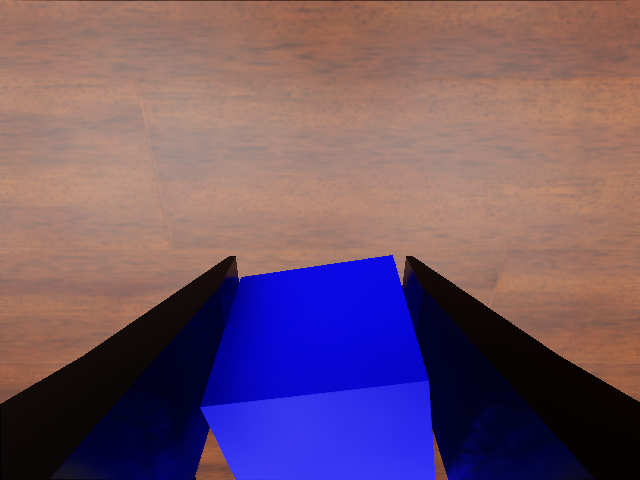}\par\vspace{3pt}
{\fontsize{8}{10}\selectfont Right wrist camera\par Observation 8, before execution~9}\end{minipage}\hfill
\begin{minipage}[t]{0.70\linewidth}\vspace{0pt}\fontsize{8.5}{10.5}\selectfont\raggedright
\textbf{Recorded decision.} Blue is visibly retained after lifting, and orange is stable at the central base. Carry blue at the tested loaded speed to the base alignment for a second-layer placement.\par
\begin{lstlisting}[style=experience,firstline=3,lastline=5]
obs, result = move_ee('right', base_xy+[1.045]+down, 0.0, max_steps=46, max_translation_step=0.008)
print(result)
print(obs['state']['right_ee_pose'])
\end{lstlisting}
\textbf{Feedback.} The loaded carry reaches its target in 41 actions with the 8 mm translation bound.
\end{minipage}\par\vspace{3pt}\noindent{\color{black!15}\rule{\linewidth}{0.35pt}}\par\vspace{3pt}\end{minipage}\par
\noindent\begin{minipage}[t]{\linewidth}
{\small\bfseries Execution 10: Place the second layer}\hfill{\footnotesize +34 actions; 282/400 used}\par\vspace{4pt}\nointerlineskip
\begin{minipage}[t]{0.27\linewidth}\vspace{0pt}\centering
\includegraphics[width=\linewidth]{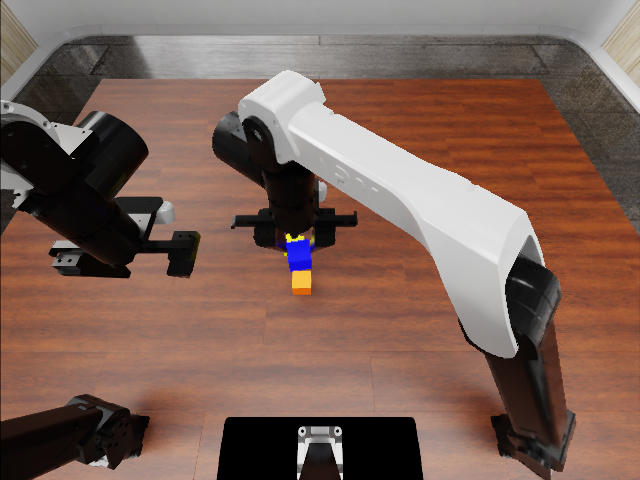}\par\vspace{3pt}
{\fontsize{8}{10}\selectfont Head camera\par Observation 9, before execution~10}\end{minipage}\hfill
\begin{minipage}[t]{0.70\linewidth}\vspace{0pt}\fontsize{8.5}{10.5}\selectfont\raggedright
\textbf{Recorded decision.} Blue remains held directly above orange, with the wrist alignment matching the calibrated reference. Lower to the second-layer release height, open, and retreat vertically to inspect the stack.\par
\begin{lstlisting}[style=experience,firstline=3,lastline=10]
obs, result = move_ee('right', base_xy+[0.968]+down, 0.0, max_steps=23, max_translation_step=0.005)
print('lower',result)
if result.get('reached',False):
    obs, result = move_ee('right', base_xy+[0.968]+down, 1.0, max_steps=8, min_steps=8)
    print('release',result)
    if not result.get('terminated',False) and not result.get('truncated',False):
        obs, result = move_ee('right', base_xy+[1.075]+down, 1.0, max_steps=10, max_translation_step=0.018)
        print('retreat',result)
\end{lstlisting}
\textbf{Feedback.} Descent, release, and retreat take 18, 8, and 8 actions. The next frame shows blue on orange.
\end{minipage}\par\vspace{3pt}\noindent{\color{black!15}\rule{\linewidth}{0.35pt}}\par\vspace{3pt}\end{minipage}\par
\noindent\begin{minipage}[t]{\linewidth}
{\small\bfseries Execution 11: Clear the right arm and approach yellow}\hfill{\footnotesize +15 actions; 297/400 used}\par\vspace{4pt}\nointerlineskip
\begin{minipage}[t]{0.27\linewidth}\vspace{0pt}\centering
\includegraphics[width=\linewidth]{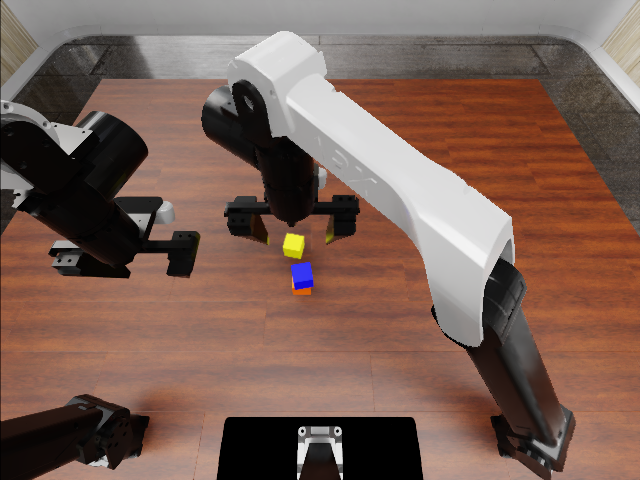}\par\vspace{3pt}
{\fontsize{8}{10}\selectfont Head camera\par Observation 10, before execution~11}\end{minipage}\hfill
\begin{minipage}[t]{0.70\linewidth}\vspace{0pt}\fontsize{8.5}{10.5}\selectfont\raggedright
\textbf{Recorded decision.} Blue is stable on orange, leaving yellow as the final layer. Move the empty right arm clear while aligning the left arm above yellow for visual calibration.\par
\begin{lstlisting}[style=experience,firstline=3,lastline=4]
obs, result = move_both([-0.05,-0.09,1.065]+down, [0.24,-0.24,1.075]+down, max_steps=20)
print(result)
\end{lstlisting}
\textbf{Feedback.} Both raised targets are reached in 15 actions; 103 native actions remain.
\end{minipage}\par\vspace{3pt}\noindent{\color{black!15}\rule{\linewidth}{0.35pt}}\par\vspace{3pt}\end{minipage}\par
\noindent\begin{minipage}[t]{\linewidth}
{\small\bfseries Execution 12: Grasp and lift yellow}\hfill{\footnotesize +50 actions; 347/400 used}\par\vspace{4pt}\nointerlineskip
\begin{minipage}[t]{0.27\linewidth}\vspace{0pt}\centering
\includegraphics[width=\linewidth]{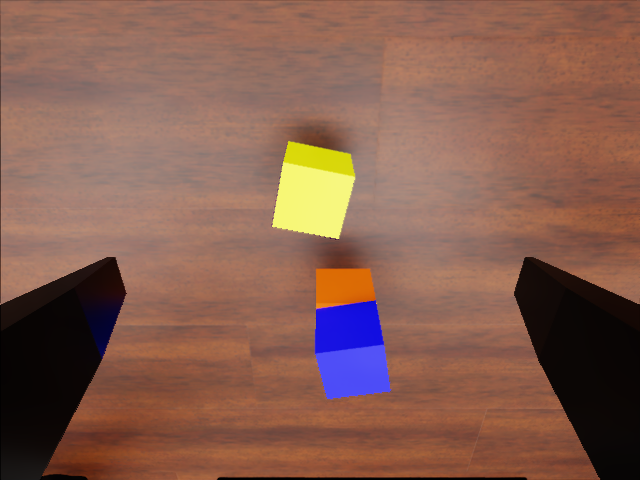}\par\vspace{3pt}
{\fontsize{8}{10}\selectfont Left wrist camera\par Observation 11, before execution~12}\end{minipage}\hfill
\begin{minipage}[t]{0.70\linewidth}\vspace{0pt}\fontsize{8.5}{10.5}\selectfont\raggedright
\textbf{Recorded decision.} Yellow is slightly left and forward of the table-height wrist alignment reference. Correct its center, grasp low, and lift vertically with clearance above the two-block stack.\par
\begin{lstlisting}[style=experience,firstline=3,lastline=8]
yellow_xy = [-0.0535,-0.110]
obs, result = move_ee('left', yellow_xy+[1.065]+down, 1.0, max_steps=5)
print('align',result)
if result.get('reached',False):
    obs, result = pick_from_above('left', yellow_xy+[0.925]+down, 1.045, action_budget=48)
    print('pick',result)
\end{lstlisting}
\textbf{Feedback.} Alignment takes 3 actions and the grasp takes 47; 53 native actions remain.
\end{minipage}\par\vspace{3pt}\noindent{\color{black!15}\rule{\linewidth}{0.35pt}}\par\vspace{3pt}\end{minipage}\par
\noindent\begin{minipage}[t]{\linewidth}
{\small\bfseries Execution 13: Carry yellow over the visible top block}\hfill{\footnotesize +12 actions; 359/400 used}\par\vspace{4pt}\nointerlineskip
\begin{minipage}[t]{0.27\linewidth}\vspace{0pt}\centering
\includegraphics[width=\linewidth]{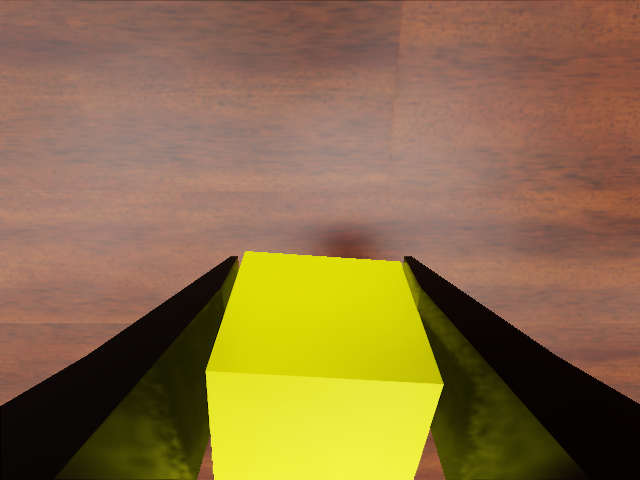}\par\vspace{3pt}
{\fontsize{8}{10}\selectfont Left wrist camera\par Observation 12, before execution~13}\end{minipage}\hfill
\begin{minipage}[t]{0.70\linewidth}\vspace{0pt}\fontsize{8.5}{10.5}\selectfont\raggedright
\textbf{Recorded decision.} Yellow is visibly retained, and the orange-blue stack remains upright. Carry yellow over the measured stack center with the tested slow translation limit.\par
\begin{lstlisting}[style=experience,firstline=3,lastline=5]
top_xy = [-0.033,-0.188]
obs, result = move_ee('left', top_xy+[1.045]+down, 0.0, max_steps=15, max_translation_step=0.008)
print(result)
\end{lstlisting}
\textbf{Feedback.} The corrected stack-center target is reached in 12 actions; 41 actions remain.
\end{minipage}\par\vspace{3pt}\noindent{\color{black!15}\rule{\linewidth}{0.35pt}}\par\vspace{3pt}\end{minipage}\par
\noindent\begin{minipage}[t]{\linewidth}
{\small\bfseries Execution 14: Place yellow and preserve a return reserve}\hfill{\footnotesize +26 actions; 385/400 used}\par\vspace{4pt}\nointerlineskip
\begin{minipage}[t]{0.27\linewidth}\vspace{0pt}\centering
\includegraphics[width=\linewidth]{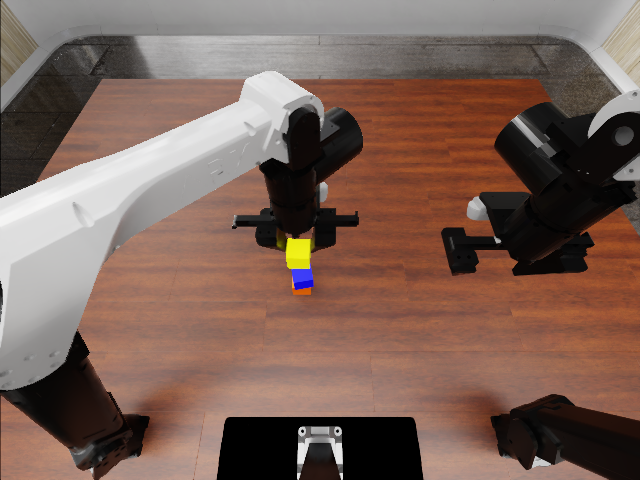}\par\vspace{3pt}
{\fontsize{8}{10}\selectfont Head camera\par Observation 13, before execution~14}\end{minipage}\hfill
\begin{minipage}[t]{0.70\linewidth}\vspace{0pt}\fontsize{8.5}{10.5}\selectfont\raggedright
\textbf{Recorded decision.} Yellow is held over the actual blue top block, with orange supporting the stack. Place the final layer slowly, release, and retreat before the home-return check.\par
\begin{lstlisting}[style=experience,firstline=3,lastline=10]
obs, result = move_ee('left', top_xy+[1.008]+down, 0.0, max_steps=13, max_translation_step=0.004)
print('lower',result)
if result.get('reached',False):
    obs, result = move_ee('left', top_xy+[1.008]+down, 1.0, max_steps=8, min_steps=8)
    print('release',result)
    if not result.get('terminated',False) and not result.get('truncated',False):
        obs, result = move_ee('left', top_xy+[1.11]+down, 1.0, max_steps=9, max_translation_step=0.018)
        print('retreat',result)
\end{lstlisting}
\textbf{Feedback.} The 4 mm descent, 8-action release, and retreat use 26 actions. Success is still false; 15 actions remain.
\end{minipage}\par\vspace{3pt}\noindent{\color{black!15}\rule{\linewidth}{0.35pt}}\par\vspace{3pt}\end{minipage}\par
\noindent\begin{minipage}[t]{\linewidth}
{\small\bfseries Execution 15: Return home and stop at official success}\hfill{\footnotesize +10 actions; 395/400 used}\par\vspace{4pt}\nointerlineskip
\begin{minipage}[t]{0.27\linewidth}\vspace{0pt}\centering
\includegraphics[width=\linewidth]{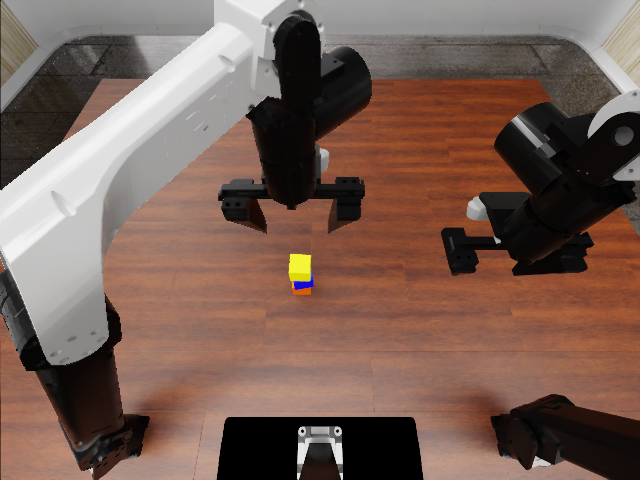}\par\vspace{3pt}
{\fontsize{8}{10}\selectfont Head camera\par Observation 14, before execution~15}\end{minipage}\hfill
\begin{minipage}[t]{0.70\linewidth}\vspace{0pt}\fontsize{8.5}{10.5}\selectfont\raggedright
\textbf{Recorded decision.} The camera shows a stable orange-blue-yellow stack after release and vertical withdrawal. Return both arms to the saved starting joints and stop as soon as the official check terminates.\par
\begin{lstlisting}[style=experience,firstline=3,lastline=8]
for i in range(15):
    obs, reward, terminated, truncated, info = step(home)
    if terminated or truncated:
        print({'steps':i+1,'reward':reward,'terminated':terminated,'truncated':truncated,'success':info.get('success',False)})
        break
print(obs['state'])
\end{lstlisting}
\textbf{Feedback.} Home commands trigger success=true and terminated=true after 10 actions. Five actions remain unused.
\end{minipage}\par\vspace{3pt}\noindent{\color{black!15}\rule{\linewidth}{0.35pt}}\par\vspace{3pt}\end{minipage}\par
\endgroup

\par\clearpage
\endgroup

\end{document}